\documentclass[11pt, a4paper, twocolumn, copyright, google]{google}

\usepackage[authoryear, sort&compress, round]{natbib}
\graphicspath{{}{./}}

\usepackage{xurl} 
\renewcommand{\today}{2026-11-13} 
\usepackage{inconsolata}  %
\usepackage{multirow}
\usepackage{tikz}
\usetikzlibrary{arrows.meta,positioning,fit,shapes.geometric,calc,backgrounds,decorations.pathreplacing}
\usepackage{pgfplots}
\pgfplotsset{compat=1.18}
\usepgfplotslibrary{groupplots}
\usepackage{listings}
\usepackage{tcolorbox}
\usepackage{array}
\newcolumntype{Y}{>{\raggedright\arraybackslash}X}
\newcommand{\resulttable}{%
  \small
  \renewcommand{\arraystretch}{1.18}%
  \setlength{\tabcolsep}{6pt}%
}
\newcommand{\maintable}{%
  \footnotesize
  \setlength{\tabcolsep}{2.5pt}%
  \renewcommand{\arraystretch}{1.12}%
}

\definecolor{cCold}{HTML}{8793A2}
\definecolor{cHT}{HTML}{C98B31}
\definecolor{cPO}{HTML}{8970AD}
\definecolor{cFull}{HTML}{087F83}
\definecolor{cSeed}{HTML}{8793A2}
\definecolor{cEvo}{HTML}{087F83}
\definecolor{crsiInk}{HTML}{203047}
\definecolor{crsiTeal}{HTML}{087F83}
\pgfplotsset{crsi axis/.style={
  axis lines*=left, axis line style={black!35,line width=.45pt},
  tick align=outside, tick style={black!35,line width=.4pt},
  tick label style={font=\sffamily\fontsize{8}{9}\selectfont,text=black!75},
  label style={font=\sffamily\fontsize{8.5}{10}\selectfont},
  title style={font=\sffamily\bfseries\fontsize{9}{10.5}\selectfont,align=left},
  grid style={solid,black!10,line width=.35pt},
  legend style={draw=none,fill=none,font=\sffamily\fontsize{8}{9}\selectfont,
    legend cell align=left,column sep=8pt},
  clip=false,
}}

\tcbuselibrary{breakable}
\newtcolorbox{casebox}[1]{breakable, colback=gray!3, colframe=gray!55, fonttitle=\small\bfseries,
  title={#1}, left=3pt, right=3pt, top=2pt, bottom=2pt, boxsep=2pt, before skip=6pt, after skip=8pt}

\uselogo{}  %
\title{Self-Improvement Memory Enables Continual Recursive Self-Improvement}
\correspondingauthor{xinyulucas@gmail.com}
\reportnumber{}  %

\makeatletter
\renewcommand\AB@affilsepx{\protect\\\protect\Affilfont}
\makeatother
\author[1,4,*]{Xin Yu}
\author[1]{Fei Xia}
\author[1]{Jiaqi Pan}
\author[1]{Youbao Tang}
\author[1]{Chengzhi Liu}
\author[2]{Adams Yu}
\author[2]{Sicong Jiang}
\author[3]{Yanfei Chen}
\author[4]{Lingzhou Xue}
\author[1]{Ting Yu}
\affil[1]{Google Cloud AI}
\affil[2]{Google DeepMind}
\affil[3]{Google Cloud AI Research}
\affil[4]{The Pennsylvania State University}
\affil[ ]{\texttt{xinyulucas@gmail.com}, \texttt{\{feixia, jiaqip, tangyoubao, chengzhiliu, adamsyuwei, jiangsicong, yanfeichen, yuti\}@google.com}, \texttt{lzxue@psu.edu}}

\newcommand{\motivationfrontpage}{}
\makeatletter
\renewcommand{\abscontent}{%
  \noindent
  \begin{minipage}[t]{\dimexpr(\textwidth-\columnsep)/2\relax}
    \vspace{0pt}%
    {\absfont\theabstract\par}
    \vspace{6pt}
    {\keywordsfont\textit{Keywords: \@keywords}\par}
  \end{minipage}\hfill
  \begin{minipage}[t]{\dimexpr(\textwidth-\columnsep)/2\relax}
    \vspace{0pt}%
    \setlength{\parskip}{0pt}
    \captionsetup{type=figure,font=small,skip=5pt}
    \centering
\includegraphics[width=\linewidth]{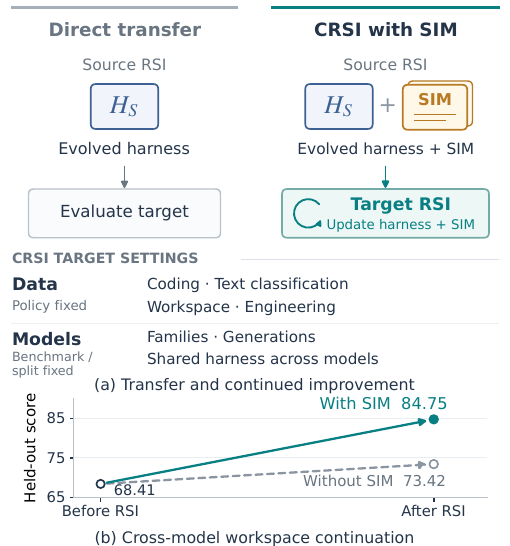}

\caption{\textbf{Continual improvement beyond direct transfer.} (a) CRSI carries the evolved harness and SIM into target RSI across data or models. SIM guides reuse and revision. (b) Both conditions continue RSI from the same Opus-evolved harness on Gemini's workspace tasks; arrows join measured endpoints with matched rounds and candidates.}
\label{fig:motivation}

  \end{minipage}%
}
\makeatother

\begin{abstract}
Harness optimization is often studied with a fixed policy model and benchmark.
When either changes, the optimizer must decide which parts of an evolved harness to reuse or revise, but the harness alone may not explain why earlier edits were made or how they performed.
In this work, we study Continual Recursive Self-Improvement (CRSI), a setting for continuing existing harness-optimization methods across compatible benchmarks or policy models.
We propose Self-Improvement Memory (SIM) for the harness optimizer, rather than the task-solving policy.
SIM groups edits by recurring harness problem and records their rationale, predicted and observed outcomes, and evaluation context.
New attempts extend this history as optimization continues.
In our experiments across four domains, SIM achieves higher final scores than harness-only and root-cause-only continuation in cross-dataset and pairwise cross-model comparisons.
Across ten independent source runs in coding, paired continuations with SIM improve final scores by an average of 2.36 points over harness-only continuation ($p=0.003$).
\end{abstract}

\begin{document}
\raggedbottom
\maketitle
{\renewcommand{\thefootnote}{*}\footnotetext{This work was done while Xin Yu was an intern at Google.}}

\section{Introduction}
\label{sec:intro}

An LLM agent depends on both its policy model and the harness that manages context, tools, and execution.
Improving the harness can raise performance without updating model weights, but finding useful changes by hand requires repeated engineering and evaluation~\citep{lin2026agentic,xia2026rrsi}.
Recent work automates this process through recursive self-improvement (RSI), using execution feedback to repeatedly propose, evaluate, and select harness edits~\citep{lee2026meta,lin2026agentic,xia2026rrsi}.
Although evolved harnesses can improve performance on other benchmarks or models~\citep{lin2026agentic,wu2026modularrsi,xia2026rrsi,xu2026adapting}, optimization is usually conducted within a fixed setting.
As models are updated and agents encounter new tasks, harness optimization often needs to continue beyond that setting.
This raises two fundamental challenges.

\ifdefined\motivationfrontpage
\else
\begin{figure}[!t]
  \input{figures/motivation_block}
\end{figure}
\fi

\textbf{Adapting to a new model.}
Model adaptation arises in two settings: \emph{multi-model serving}, where one shared harness supports different policy models, and \emph{model evolution}, where new versions replace earlier ones within a model family.
Both motivate reusing an evolved harness and its improvement experience instead of independently optimizing a harness for every model or release.
Yet modifications suited to one policy can interact differently with another, even when the benchmark stays fixed~\citep{yang2026better,yu2026co}.
In our workspace experiment, an Opus-evolved harness starts at 68.41 on Gemini, below the seed harness's 73.89 (Figure~\ref{fig:crossmodel}).
The optimizer must determine which inherited mechanisms still help and which need revision.

\textbf{Adapting to new data.}
With the policy fixed, a different task distribution can expose failures that source optimization did not address.
A completion gate that checks whether commands ran successfully, for example, may need to check whether a feasible solution is also optimal.
In our text-classification experiment, the source-evolved FewShotAll harness starts at 72.92 on the target suite, below the seed's 75.72 (Figure~\ref{fig:main}).
A useful source fix may therefore need revision before it helps the target.

In either transition, the evolved harness alone leaves important questions unanswered: why was an edit proposed, what was it expected to fix, and what happened when it was tested?
Source optimization produces this evidence, but it need not accompany the harness.
Without this history, target optimization may repeat earlier attempts or change inherited components without knowing their purpose.
We therefore ask whether the experience gained during source optimization can help the harness adapt to new models or data.

In this work, we study \textbf{Continual Recursive Self-Improvement (CRSI)}, in which harness optimization continues after a benchmark or policy-model change while preserving a compatible executable interface.
We propose \textbf{Self-Improvement Memory (SIM)} to carry improvement experience alongside the evolved harness (Figure~\ref{fig:motivation}).
SIM groups selected edits by the harness problems they address and gives the proposer and selector access to their rationale, predicted and observed outcomes, and evaluation context.
It creates entries for new problems and appends further attempts and outcomes to existing ones, helping the optimizer decide which modifications to retain, revise, replace, or remove.

Our contributions are as follows:
\begin{itemize}[leftmargin=*, itemsep=2pt, topsep=3pt]
  \item \textbf{Continual harness optimization.} We define cross-dataset continuation with the policy fixed and cross-model continuation with the benchmark and split fixed. Existing harness-optimization methods can use this setting while retaining their proposal, evaluation, and selection procedures.
  \item \textbf{Self-Improvement Memory.} We introduce a structured record of harness problems, modifications, and outcomes that conditions both proposal and selection, preserving source experience while incorporating new attempts during target optimization.
  \item \textbf{Controlled evidence across four domains.} With matched starting harnesses, round counts, and candidate budgets, Full SIM outperforms harness-only and root-cause-only continuation in six cross-dataset and four Opus-to-Gemini settings. Across ten independent source runs in coding, paired continuations with SIM improve 9 of 10 runs, with a mean advantage of 2.36 points over harness-only continuation ($p=0.003$).
\end{itemize}

\section{Related Work}
\label{sec:related}

\paragraph{Recursive self-improvement of agent harnesses.}
\citet{yu2026harnesssurvey} survey harness optimization workflows, recursive reuse, and the conditions under which evolved modifications transfer.
Meta-Harness and AHE search over harness code using execution traces and scores~\citep{lee2026meta,lin2026agentic}.
Other approaches let the agent modify its own system~\citep{zhang2026self,zhang2025darwin,zhang2026hyperagents} or learn a policy for editing harnesses~\citep{shao2026harness}.
SIA instead combines harness edits with updates to the task model's weights~\citep{hebbar2026sia}.

\paragraph{Transfer of evolved harnesses.}
Evolved harnesses can retain their gains on held-out tasks, other models, or downstream benchmarks~\citep{lin2026agentic,lee2026meta,shi2026pluginrsi}.
Recent methods encourage transfer through benchmark-disjoint evolution and modular edits~\citep{wu2026modularrsi}, regularized search~\citep{xia2026rrsi}, or separation of general and benchmark-specific mechanisms~\citep{yu2026harmony}.
Their effectiveness still depends on how the model and harness interact~\citep{yang2026better,yu2026co}.
PluginRSI also reuses its plugin library in subsequent optimization.

\paragraph{Memory of the improvement process.}
Prior work retains candidate artifacts~\citep{lee2026meta}, edit histories~\citep{xia2026rrsi}, improvement graphs~\citep{zhang2026causal}, skill-revision records~\citep{li2026skillhone}, and playbooks~\citep{zhang2026turbo}.
HarnessBank groups candidate harnesses by the problems they address~\citep{luo2026harnessbank}.
\citet{xu2026self} continues multi-task evolution with free-text notes and source replay; Harness$^2$ reuses an improvement library across benchmarks~\citep{xu2026harness}.
Meta-Skill and Meta$^2$ transfer learned guidance for constructing or editing task-specific harnesses~\citep{qian2026learning,yuan2026meta}.
Continual adaptation also raises retention and interference concerns~\citep{chen2026continual,zhao2026coding}.
SIM links harness problems to modifications, predicted and observed outcomes, and provenance.
We test SIM by continuing optimization from the same evolved harness with no SIM, root causes only, or the full record after a benchmark or model change.
Appendix~\ref{sec:app-related} expands these comparisons, including continued optimization and memory for task-solving agents.

\section{Method}
\label{sec:method}

\begin{figure*}[t]
  \centering
  \input{figures/overview_include}
  \caption{\textbf{SIM in the agentic outer loop.} Starting from $(H_S,\mathcal{M}_S)$, the proposer and selector consult SIM (dashed arrows) to improve the harness on evolve-set tasks. After selection, SIM records the chosen edit and its measured outcome (teal arrow). Held-out evaluation stays outside the loop.}
  \label{fig:overview}
\end{figure*}

\subsection{Problem Formulation}
\label{sec:method-problem}

\paragraph{Harness RSI.}
Following Meta-Harness~\citep{lee2026meta}, we optimize an executable harness $H$ around a fixed policy model $M$.
The harness determines the model's context, tools, and execution flow.
We write a setting as $\sigma=(M,\mathcal{X})$, with disjoint evolve and held-out evaluation sets $\mathcal{X}^{\mathrm{evo}}$ and $\mathcal{X}^{\mathrm{eval}}$.
The score $J(H;M,\mathcal{X})$ follows the benchmark's aggregation rule (Appendix~\ref{sec:app-benchmarks}).
Only evolve-set feedback is available for proposing edits, selecting candidates, updating SIM, or deciding when to stop.

A run consists of $N$ rounds.
At round $t=0,\ldots,N-1$, the evaluator executes $H_t$ on $\mathcal{X}^{\mathrm{evo}}$ and returns scores, traces, and errors as $F_t$.
The proposer $P$ uses $F_t$ and the run's history $\mathcal{D}_t$ (earlier feedback and code diffs) to generate $k$ candidate harnesses $\mathcal{C}_t$.
The selector $\mathcal{S}$ screens proposals before execution and chooses among the evaluated candidates, weighing performance, complexity, domain specificity, and regression risk:
\begin{equation}
  \begin{aligned}
    \mathcal{C}_t &= P(H_t,F_t,\mathcal{D}_t), \\
    H_{t+1} &= \mathcal{S}(H_t,F_t,\mathcal{C}_t,F_t^{\mathcal{C}}),
  \end{aligned}
  \label{eq:rsi}
\end{equation}
Here $F_t^{\mathcal{C}}$ contains feedback from the candidates that passed screening and were executed.
If no candidate is accepted, $H_{t+1}=H_t$.
The proposer and selector are LLM-based harness engineers; $M$ is the policy that executes the tasks.

\paragraph{Continual setting.}
After source RSI in $\sigma_S=(M_S,\mathcal{X}_S)$, optimization resumes from the final retained harness $H_S$ in $\sigma_T=(M_T,\mathcal{X}_T)$.
In \emph{cross-dataset} CRSI, the benchmark changes while the policy stays fixed ($\mathcal{X}_S\neq\mathcal{X}_T$, $M_S=M_T$).
In \emph{cross-model} CRSI, the policy changes on the same benchmark and split ($M_S\neq M_T$, $\mathcal{X}_S=\mathcal{X}_T$).
Both require a shared executable harness interface.
We measure target improvement as the final minus starting score under $J(\cdot\,;M_T,\mathcal{X}_T^{\mathrm{eval}})$.

\subsection{Self-Improvement Memory}
\label{sec:method-sim}

SIM groups edits by the recurring harness problems they address.
For a root-cause hypothesis $c_i$, such as premature completion, an entry is
\begin{equation}
  m_i=\big(c_i,[(a_{ij},o_{ij},p_{ij})]_j\big),
  \label{eq:sim-entry}
\end{equation}
where $j$ indexes attempts in order and $\mathcal{M}_t=\{m_i\}$ is the current memory.
Each attempt stores the modification and its rationale $a_{ij}$, predicted and observed task outcomes $o_{ij}$, and provenance $p_{ij}$ (domain, benchmark, policy, and round).
Predictions of which tasks an edit will fix are recorded before evaluation and checked afterward.

At each round, the proposer receives the complete SIM in its system prompt, and the selector reads the same record when assessing candidates.
Both read all entries without retrieval or truncation.
The loop in Equation~\ref{eq:rsi} becomes
\begin{equation}
  \begin{aligned}
    \mathcal{C}_t &= P(H_t,F_t,\mathcal{D}_t;\mathcal{M}_t), \\
    H_{t+1} &= \mathcal{S}(H_t,F_t,\mathcal{C}_t,F_t^{\mathcal{C}};\mathcal{M}_t).
  \end{aligned}
  \label{eq:rsi-sim}
\end{equation}

Each proposal cites an existing root cause or names a new one.
After selection, \emph{only the chosen candidate} is recorded in $e_t$, including its measured outcomes even when the edit regresses or fails to fix a predicted task.
The update $\mathcal{M}_{t+1}=U(\mathcal{M}_t,e_t)$ appends this record to the cited problem or creates a new entry, preserving earlier attempts.
If no edit is accepted, $e_t=\emptyset$ and $U(\mathcal{M}_t,\emptyset)=\mathcal{M}_t$.
Entries with identical root-cause text are merged without dropping their histories.
Appendix~\ref{sec:app-sim-interfaces} gives the entry-matching details.

\subsection{Continual RSI via SIM}
\label{sec:method-crsi}

Source RSI starts with an empty SIM and produces $(H_S,\mathcal{M}_S)$.
Full SIM carries both into target optimization (Figure~\ref{fig:overview}).
The proposer and selector use each edit's rationale, outcome, and evaluation context to judge whether it still applies to the current failures.
They can retain, revise, replace, or remove harness modifications while keeping the earlier memory records.
Appendix~\ref{sec:app-sim-rules} gives the prompt instructions for using this history.

\paragraph{Transfer conditions.}
We compare four conditions: \emph{Cold Start} $(H_0,\emptyset)$ from the seed harness; \emph{Harness Transfer} $(H_S,\emptyset)$; \emph{Problem-Only Memory} $(H_S,\mathcal{M}^{c}_S)$ with root causes only; and \emph{Full SIM} $(H_S,\mathcal{M}_S)$.
The three transfer conditions start from the same $H_S$, and all four use the same target policy, tasks, and round and candidate budgets.
Each keeps its assigned memory condition throughout target RSI, while retaining the ordinary target-run history $\mathcal{D}_t$ (Appendix~\ref{sec:app-transfer-conditions}).
Source history enters through SIM; new target records follow the same memory condition.

\section{Experiments}
\label{sec:experiments}

Our experiments address four questions:
\begin{itemize}[label=$\bullet$, leftmargin=*, itemsep=1pt, topsep=3pt]
  \item Does SIM improve continued RSI after a benchmark or policy-model change? (\S\ref{sec:results}--\ref{sec:results-crossmodel})
  \item Does the full SIM record help beyond retaining only root causes? (\S\ref{sec:results} and \S\ref{sec:results-model-transfer})
  \item How does SIM guide the reuse and revision of inherited harness edits? (\S\ref{sec:analysis}; Appendix~\ref{sec:app-transfer})
  \item Is SIM's benefit consistent across independently evolved source harnesses? (\S\ref{sec:analysis})
\end{itemize}

\subsection{Experimental Setup}
\label{sec:setup}

\paragraph{Domains.}
We study four domains in which a source-evolved harness remains executable on the target (Appendix~\ref{sec:app-benchmarks}, Table~\ref{tab:benchmarks}).
\emph{Coding} moves from Terminal-Bench 2.1 (TB2.1) to Terminal-Bench 4 (TB4)~\citep{merrill2026terminal,tbench2026v21,tbench2026v4}, containerized shell tasks scored by their own tests.
\emph{Text classification} uses the online classification environment of Meta-Harness~\citep{lee2026meta}, from a three-dataset source suite to an eight-dataset target suite absent from source RSI.
\emph{Agentic workspace} moves from Harvey LAB~\citep{harvey2026lab} to the equal-weight mean of JobBench~\citep{li2026jobbench}, GDPval~\citep{patwardhan2026gdpval}, and APEX-Agents~\citep{vidgen2026apex}, using scoring interfaces adapted from RRSI~\citep{xia2026rrsi}, with Gemini 3.8 Flash as the judge.
\emph{Engineering design} moves from EngDesign~\citep{guo2025engdesign} to Frontier-Eng~\citep{chi2026frontiereng}, scored by frozen verifiers.
Tasks are split uniformly at random, without stratification, into evolve and held-out sets of approximately equal size; exact counts are in Appendix~\ref{sec:app-benchmarks}.
Only evolve-set scores and trajectories reach the proposer, selector, and SIM.
The main comparisons use five rollouts per task, with every rollout included in the score.

\paragraph{Transitions, models, and budgets.}
Cross-dataset CRSI uses Gemini 3.8 Flash~\citep{google2026gemini38} as the policy in both stages.
For the cross-model transfer comparison, source RSI uses Claude Opus 5.5~\citep{anthropic2026opus55}, then Gemini 3.8 Flash continues on the same benchmark and split.
Both stages use the target benchmarks listed above: TB4, the eight-dataset classification suite, the merged workspace benchmarks, and Frontier-Eng.
Policy weights remain frozen.
Gemini 3.8 Flash also serves as proposer and selector in the main comparisons.
A source-RSI ablation replaces it with Claude Opus 5.5 in these roles, yielding larger gains from both coding seeds (Appendix~\ref{sec:app-harness-agent}, Table~\ref{tab:harness-agent}).
The seeds are the minimal AHE seed~\citep{lin2026agentic} and Terminus-2~\citep{merrill2026terminal} for coding, NoMemory and FewShotAll~\citep{lee2026meta} for text classification, and a ReAct agent~\citep{yao2022react} over an MCP tool gateway~\citep{vidgen2026apex} for workspace and engineering.
For the latter two domains, we adopt RRSI's environments, seed harnesses, and evaluation interfaces~\citep{xia2026rrsi}; Appendix~\ref{sec:app-setup} details our experimental setup.
Every source and target run uses $N=10$ rounds with $k=5$ candidate proposals per round in the proposer--selector loop described in \S\ref{sec:method-problem}.
Proposals rejected before execution receive no evaluation score; the five rollouts per task specify the separate evaluation allocation.

\paragraph{Comparison conditions.}
Cold Start begins from the seed harness without SIM.
Harness Transfer, Problem-Only Memory, and Full SIM share the same source-evolved harness, with no SIM, root causes only, and the complete record, respectively.
These memory conditions persist throughout target optimization, with matched target tasks, policy, round counts, and candidate budgets (\S\ref{sec:method-crsi}).

\paragraph{Metrics and reporting.}
Scores are on a 0--100 scale to two decimals: pass@1 for coding, accuracy pooled over examples for text classification, benchmark-specific rubric, win-rate, and pass@1 scores for workspace, and pass rate or medal score for engineering (Appendix~\ref{sec:app-benchmarks}).
Before and after scores evaluate the starting harness and the version retained after the final target round on the held-out set.
Except for the dedicated paired study, each condition is one complete RSI run, and its five rollouts per task estimate execution performance rather than optimization variance.
We report individual settings descriptively and use the ten-pair coding study to assess variation across independent source runs (\S\ref{sec:analysis}).

\subsection{Cross-Dataset CRSI}
\label{sec:results}

\begin{figure*}[!t]
  \centering
  \input{figures/main_results_include}
  \caption{Cross-dataset CRSI: held-out target scores before (open bars) and after (filled bars) target RSI. The three transfer conditions share the source-evolved harness; Cold Start uses the seed. Axis limits vary by domain. Table~\ref{tab:main-results} lists exact scores.}
  \label{fig:main}
\end{figure*}

\paragraph{SIM improves final scores in all six dataset transitions.}
The memory ablation in Figure~\ref{fig:main} compares continued optimization from the same evolved harness with no SIM, root causes only, or the full record.
Full SIM finishes highest in all six settings, exceeding Harness Transfer by 2.43--4.35 points and Problem-Only Memory by 0.20--6.06 points.
Problem-Only Memory has mixed effects: it ranges from 1.82 points below Harness Transfer to 2.40 above it.
Keeping the edit history and evaluation evidence thus helps beyond retaining the diagnosis alone (Appendix~\ref{sec:app-memory-ablation}).

The margin over the strongest alternative ranges from 0.20 points on NoMemory to 4.35 on engineering.
Section~\ref{sec:analysis} tests whether the benefit recurs across independently evolved source harnesses.
Appendix~\ref{sec:app-bars} lists the exact starting and final scores for both transfer settings.

\paragraph{Source improvement does not ensure a better target start.}
Source RSI raises held-out source scores by 2.75--6.41 points in all six settings (Appendix~\ref{sec:app-source}), yet the FewShotAll harness starts target RSI below the seed: 72.92 versus 75.72.
Harness-only continuation reaches 74.12 and remains below Cold Start at 75.92, while Full SIM reaches 77.92.

On NoMemory, Cold Start gains more than Full SIM (3.40 versus 2.20 points) but finishes lower (72.00 versus 73.60).
We therefore report both endpoints and use Harness Transfer, which shares Full SIM's initialization, as the direct control for the memory procedure.

\subsection{Cross-Model CRSI}
\label{sec:results-crossmodel}

We examine transfer between model families, continuation across model generations, and a harness shared by multiple models.

\subsubsection{Harness Transfer across Model Families}
\label{sec:results-model-transfer}

\begin{figure*}[!t]
  \centering
  \input{figures/cross_model_results_include}
  \caption{Cross-model CRSI from Opus 5.5 to Gemini 3.8 Flash, with benchmark and split fixed. Open and filled bars show held-out scores before and after Gemini's target RSI. Transfer conditions share the Opus-evolved harness; Cold Start uses the seed (Table~\ref{tab:crossmodel}).}
  \label{fig:crossmodel}
\end{figure*}

When the Opus-evolved harness is first run with Gemini, it scores below Gemini's seed harness in every domain (Figure~\ref{fig:crossmodel}).
The deficits range from 0.60 points in text classification to 5.48 in workspace.
Target RSI does not always overcome this disadvantage: Harness Transfer still finishes below Cold Start in three domains.
Full SIM finishes highest in all four and exceeds harness-only continuation by 3.60--11.33 points.
It also finishes above Problem-Only Memory in all four domains (Appendix~\ref{sec:app-memory-ablation}).
Workspace shows the largest difference: both begin at 68.41, but Full SIM reaches 84.75 while Harness Transfer reaches 73.42.

\subsubsection{Continual Improvement across Model Generations}
\label{sec:results-generations}

\begin{table}[t]
  \centering
  \maintable
  \begin{tabular*}{\columnwidth}{@{}l@{\extracolsep{\fill}}rrr@{}}
    \toprule
    \textbf{Gemini model} & \textbf{Seed} & \textbf{Fresh RSI} & \textbf{CRSI} \\
    \midrule
    2.5 Flash & 17.8 & 21.4 & \textbf{22.4} \\
    3.5 Flash & 70.2 & 73.3 & \textbf{76.8} \\
    3.8 Flash & 87.2 & 88.9 & \textbf{90.5} \\
    4 Argon & 89.7 & \textbf{95.6} & 94.8 \\
    \bottomrule
  \end{tabular*}
  \caption{Success rates on held-out TB2.1 across Gemini generations. Seed uses the unevolved harness; Fresh RSI optimizes it independently. CRSI adds SIM in the first generation and inherits the previous generation's harness and SIM thereafter. Bold compares Fresh RSI and CRSI.}
  \label{tab:generations}
\end{table}

The generation chain carries the evolved harness and SIM through successive Gemini models on TB2.1 (Table~\ref{tab:generations}; setup in Appendix~\ref{sec:app-generations}).
Fresh RSI starts from the seed at each generation.
In the first generation, both procedures use the same seed, so that comparison tests SIM during the initial run.
\emph{Continuation improves over fresh RSI in two of the three subsequent generations}: by 3.5 points on Gemini 3.5 Flash and 1.6 on Gemini 3.8 Flash, while finishing 0.8 points lower on Gemini 4 Argon.

\subsubsection{One Harness for Multiple Models}
\label{sec:results-shared}

\begin{table}[t]
  \centering
  \maintable
  \begin{tabular*}{\columnwidth}{@{}l@{\extracolsep{\fill}}rrr@{}}
    \toprule
    \textbf{Harness construction} & \textbf{Evolve} & \textbf{Held-out} & \textbf{All} \\
    \midrule
    B1: Per-model harnesses & 77.40 & 75.20 & 76.30 \\
    B2: Shared seed & 72.36 & 73.28 & 72.82 \\
    \addlinespace[3pt]
    S1: Sequential CRSI & 76.72 & 74.20 & 75.46 \\
    S2: Unified synthesis & \textbf{78.92} & \textbf{75.24} & \textbf{77.08} \\
    \midrule
    \multicolumn{4}{@{}l}{\emph{S3: Source policy used for RSI}} \\
    \quad Gemini 3.8 Flash & 74.28 & 71.24 & 72.76 \\
    \quad Grok 4.20 Reasoning & 68.24 & 66.84 & 67.54 \\
    \quad Claude Opus 5.5 & 75.20 & 72.24 & 73.72 \\
    \bottomrule
  \end{tabular*}
  \caption{Text-classification scores averaged over Grok 4.20 Reasoning, Claude Opus 5.5, and Gemini 3.8 Flash. S3 rows name the policy used to evolve each harness; each harness is evaluated with all three models. All averages the equally sized evolve and held-out splits. Bold marks the highest mean in each column.}
  \label{tab:multimodel}
\end{table}

We also ask whether one harness can serve Grok, Opus, and Gemini on the text-classification target suite (Table~\ref{tab:multimodel}; setup in Appendix~\ref{sec:app-shared}).
We compare sequential CRSI, unified synthesis using feedback from all three models, and transfer of harnesses evolved separately with each policy.
\emph{Unified synthesis reaches a similar mean score to separately evolved harnesses}: 75.24 versus 75.20 on held-out examples, above the shared seed's 73.28.
Sequential CRSI reaches 74.20; the three S3 harnesses average 66.84--72.24 across the evaluation models.
This descriptive comparison assesses a common harness; the matched starting-harness comparisons in \S\ref{sec:results-model-transfer} test the full SIM procedure.

\subsection{Analysis: The Role of SIM in CRSI}
\label{sec:analysis}

The paired comparisons test whether SIM improves continuation from a common starting harness.
We then follow individual memory entries to see how earlier diagnoses inform later edits.

\begin{figure*}[!t]
  \centering
  \input{figures/convergence_include}
  \caption{Coding optimization (TB2.1$\rightarrow$TB4). (a,b)~Held-out TB4 scores of retained harnesses after each target round; round 0 is the starting harness. (c)~Final-score differences (SIM minus No SIM) across ten paired continuations from independently evolved Terminus-2 harnesses. The dashed line marks the mean; Table~\ref{tab:paired} gives individual scores.}
  \label{fig:convergence}
\end{figure*}

\begin{table}[!t]
  \centering
  \maintable
  \setlength{\tabcolsep}{4pt}
  \begin{tabular*}{\linewidth}{@{}r@{\extracolsep{\fill}}rrrr@{}}
    \toprule
    \textbf{Run} & \textbf{Start} & \textbf{No SIM} & \textbf{SIM} & $\Delta$ \\
    \midrule
    1 & 18.18 & 20.61 & 22.42 & $+1.81$ \\
    2 & 16.97 & 17.58 & 19.39 & $+1.81$ \\
    3 & 17.58 & 20.00 & 20.61 & $+0.61$ \\
    4 & 18.18 & 18.18 & 23.64 & $+5.46$ \\
    5 & 17.58 & 18.79 & 20.61 & $+1.82$ \\
    6 & 19.39 & 20.00 & 23.64 & $+3.64$ \\
    7 & 18.79 & 20.00 & 21.82 & $+1.82$ \\
    8 & 18.18 & 20.00 & 19.39 & $-0.61$ \\
    9 & 19.39 & 18.18 & 23.03 & $+4.85$ \\
    10 & 19.39 & 20.00 & 22.42 & $+2.42$ \\
    \midrule
    \textbf{Mean} & 18.36 & 19.33 & \textbf{21.70} & $\mathbf{+2.36}$ \\
    \bottomrule
  \end{tabular*}
  \caption{Paired coding continuations from ten independent Terminus-2 source runs. Start is the evolved harness's held-out TB4 score; No SIM and SIM are scores after ten target rounds. $\Delta$ is SIM minus No SIM (mean $+2.36$; paired $p=0.003$).}
  \label{tab:paired}
\end{table}

\paragraph{SIM improves continuation from the same evolved harness.}
We repeat source RSI ten times from Terminus-2 on TB2.1 and initialize two ten-round TB4 runs from each resulting harness.
Within each pair, the starting harness, policy, benchmark, and round and candidate budgets match.
One run inherits that source run's SIM and appends target attempts; the other continues without SIM.
The independent source run is the unit of analysis.

SIM finishes higher in 9 of 10 pairs (Table~\ref{tab:paired}), with one pair favoring continuation without SIM by 0.61 points.
From a shared mean starting score of 18.36, the mean final score is 21.70 with SIM and 19.33 without it.
The mean paired advantage is 2.36 points (95\% CI $[1.05,3.68]$; two-sided paired $t(9)=4.06$, $p=0.003$).
A two-sided signed-rank check gives $p=0.006$; the paired standardized effect is $d_z=1.28$ (statistical details in Appendix~\ref{sec:app-repetitions}).

\paragraph{SIM guides reuse and revision of inherited edits.}
Full SIM and Harness Transfer inherit the same completion gate from Terminus-2 source RSI (Figure~\ref{fig:with-without}b, Appendix~\ref{sec:app-component}).
With SIM, the proposer uses the gate's rationale and source outcomes to retain completion verification while adding target checks, including feasibility versus optimality.
Without SIM, the run tightens the gate, twice repairs completion deadlocks, and misattributes the inherited gate to its own first round.
The requirement carries over; its implementation needs revision.
Appendices~\ref{sec:app-transfer} and~\ref{sec:app-case-lessons} summarize transfer patterns across domains and models and the lessons for SIM.

\paragraph{SIM preserves diagnoses while adding target evidence.}
P3 records a mismatch between deliverables and the verifier's interface contract (Figure~\ref{fig:sim-p3}; Appendix~\ref{sec:app-case-p3}).
Target rounds 5 and 6 append revised checks and measured outcomes under this diagnosis, keeping source attempts and unresolved predictions available to later rounds.

\paragraph{SIM supports earlier target progress and source retention in coding.}
Full SIM is best or tied throughout rounds 1--10 for both coding seeds (Figure~\ref{fig:convergence}a,b).
It exceeds the strongest control's final score (20.61) by round 5 from AHE and round 2 from Terminus-2.
In the AHE source-retention trace, all transfer conditions start at 90.22 on held-out TB2.1 (Figure~\ref{fig:forgetting}, Appendix~\ref{sec:app-retention}).
Full SIM loses 3 successful rollouts out of 225, versus 4 for Problem-Only Memory and 6 for Harness Transfer, while gaining more on TB4 than Problem-Only Memory (6.67 versus 0.61 points).

\section{Conclusion}
\label{sec:conclusion}

CRSI continues harness optimization when the benchmark or policy model changes.
SIM carries earlier diagnoses, edits, and outcomes into this process and records new attempts.
From matched starting harnesses, Full SIM outperforms harness-only continuation in all six cross-dataset settings and four Opus-to-Gemini domains.
The ten-pair coding study finds a mean advantage of 2.36 points across independent source runs.
The cases show how this history helps the proposer and selector reuse earlier reasoning while revising fixes for new requirements.

\section*{Limitations}
\label{sec:limitations}

Within our compute budget, independent RSI repetitions focus on coding; the other domains are evaluated with one run per condition.
We study settings with compatible harness interfaces and keep the optimizer's planning and search procedures fixed.
Our comparisons evaluate SIM as a complete procedure, combining inherited memory with updates during target optimization.

Future work can extend the evaluation to more repeated runs and longer transition sequences, and explore adaptation of planning, token allocation, and tool interfaces.

\section*{Ethics Statement}
\label{sec:ethics}

LLM-based optimization agents rewrite harness code in our experiments.
The resulting harnesses run inside the benchmarks' execution environments, with policy weights fixed and scores computed by frozen verifiers or a fixed judging procedure.
A self-modifying harness could satisfy an evaluator without solving the intended task.
We keep held-out feedback away from the optimization agents, identify each benchmark's verifier or judge, and include an accepted edit that failed to repair its target tasks (Appendix~\ref{sec:app-case-gating}).
Workspace scores may also reflect biases of the LLM judge.
The experiments use existing publicly available benchmarks and datasets.

\setlength{\bibsep}{5pt plus 1pt minus 1pt} 
\bibliography{custom}

\appendix
\clearpage
\raggedbottom
\section*{Guide to the Appendices}
Appendices~\ref{sec:app-benchmarks}--\ref{sec:appendix} specify evaluation, experimental setup, and SIM implementation.
Appendix~\ref{sec:app-results} gives additional numerical results and ablations; Appendix~\ref{sec:app-cases} explains the observed edits through case studies.

\begin{center}
\small
\renewcommand{\arraystretch}{1.18}
\begin{tabularx}{\linewidth}{@{}lX@{}}
\toprule
\textbf{Appendix} & \textbf{Contents} \\
\midrule
\ref{sec:app-related} & \textbf{Extended Related Work}: harness optimization, transfer, and memory. \\
\ref{sec:app-benchmarks} & \textbf{Benchmarks and Evaluation}: datasets, splits, feedback access, and scoring. \\
\ref{sec:app-setup} & \textbf{Experimental Setup}: models, seeds, budgets, comparison conditions, and repeated runs. \\
\ref{sec:appendix} & \textbf{SIM Implementation}: how agents read, update, and store SIM. \\
\ref{sec:app-results} & \textbf{Additional Results}: source and target scores, source retention, and ablations. \\
\ref{sec:app-cases} & \textbf{Case Studies}: source RSI, continual adaptation, memory records, and transfer patterns. \\
\bottomrule
\end{tabularx}
\end{center}

\section{Extended Related Work}
\label{sec:app-related}

This appendix expands \S\ref{sec:related} around three questions: how harnesses are improved, how optimization continues after a change of setting, and what role memory plays in that process.

\subsection{How Are Agent Harnesses Improved?}
\label{sec:app-related-optimization}

Harness RSI optimizes prompts, tools, control flow, memory, and context management around a policy model.
GEPA uses trajectory reflection and recombination to optimize prompts~\citep{agrawal2026gepa}.
Meta-Harness exposes prior implementations, execution traces, and scores through a shared filesystem~\citep{lee2026meta}.
AHE makes harness components and modification outcomes observable, checking edit-level predictions against subsequent task outcomes~\citep{lin2026agentic}.
In ML development, MLToolBench learns diagnostic tool use through supervised and reinforcement learning~\citep{yu2026mltoolbench}.

Other methods change how improvement itself is performed.
Self-Harness combines weakness mining, harness proposals, and regression-based validation~\citep{zhang2026self}.
DGM and Hyperagents explore self-modification and open-ended evolution of agent programs~\citep{zhang2025darwin,zhang2026hyperagents}; Hyperagents also reports transfer of meta-level improvements across domains.
Harness-R1 learns a harness-editing policy from failure trajectories using reinforcement learning~\citep{shao2026harness}, while SIA updates both the harness and the task model's weights~\citep{hebbar2026sia}.
\citet{bai2026verify} train idea-level critics to screen ML modifications before costly empirical verification.
MLToolBench learns task-agent behavior, while the critics learn candidate assessment; SIM preserves edit histories for the proposer and selector.
CRSI defines continuation after a benchmark or policy change for compatible proposal--evaluation--selection loops (\S\ref{sec:method-problem}).

\subsection{How Does Optimization Continue after Transfer?}
\label{sec:app-related-transfer}

\paragraph{Transfer of evolved harnesses.}
AHE evaluates transfer from Terminal-Bench to SWE-bench Verified without re-evolving the harness~\citep{lin2026agentic}; Meta-Harness and PluginRSI report transfer across solver models~\citep{lee2026meta,shi2026pluginrsi}.
Several methods aim to improve this transfer.
ModularRSI uses benchmark-disjoint evolution data, contrasts successful and failed trajectories, and directs edits toward recurring deficiencies in functional modules~\citep{wu2026modularrsi}.
RRSI constrains bundled modifications and redirects stalled search toward underexplored components~\citep{xia2026rrsi}.
Harmony Harness separates general mechanisms from benchmark-specific adaptations~\citep{yu2026harmony}.
The interaction between a model and its harness can nevertheless require model-specific adaptation~\citep{yang2026better,yu2026co}.

\paragraph{Continued optimization.}
Prior work also reuses improvement experience during further optimization.
PluginRSI reuses an evolved plugin library when optimizing from the initial harness~\citep{shi2026pluginrsi}.
\citet{xu2026self} continues multi-task harness evolution on a new benchmark with free-text evolution notes and source-task replay.
Harness$^2$ retains adopted harnesses and refinement lessons in a library and tests its reuse across benchmarks~\citep{xu2026harness}.
Related settings include task streams and continuous episodes~\citep{karten2026continual,wei2026evo,xiao2026recurrent}.
Adaptive Auto-Harness separates a persistent evolution workspace from deployable harnesses, with optional routing among stored branches~\citep{liu2026adaptive}.
EvoHarnessBench studies retention and adaptation as externally supplied tools, skills, and specialist agents expand~\citep{ke2026evoharnessbench}.

\paragraph{Evaluating continuation.}
Held-out evaluation and baselines with matched feedback and inference budgets help distinguish improvements from additional optimization resources~\citep{wang2026rethinking}.
SEAGym tracks validation, transfer, replay, and cost throughout evolution~\citep{zheng2026seagym}.
Studies of shifting task streams and coding benchmarks also document interference and forgetting~\citep{chen2026continual,zhao2026coding}.
We report the evolved harness's score before and after target RSI, compare memory conditions from the same starting harness with matched round and candidate budgets, and separately track source retention in one coding transition (Appendix~\ref{sec:app-retention}).

\subsection{What Does Memory Store, and Who Uses It?}
\label{sec:app-related-memory}

\paragraph{Memory for harness optimization.}
Keeping improvement history has several precedents.
Meta-Harness retains candidate artifacts~\citep{lee2026meta}; RRSI records hypotheses, diffs, score and cost changes, and acceptance decisions, including rejected edits~\citep{xia2026rrsi}.
The Causal Improvement Graph links evidence, hypotheses, interventions, and outcomes~\citep{zhang2026causal}.
SkillHone preserves diagnoses, revisions, and evaluation evidence~\citep{li2026skillhone}, and Turbo Harness summarizes completed optimization into a playbook for instance-specific edits~\citep{zhang2026turbo}.
HarnessBank groups candidates by modified component and diagnosed failure, retaining high-performing harnesses for recombination~\citep{luo2026harnessbank}.
Organizing improvement around problems therefore also has precedent.
Meta-Skill freezes learned support-design principles to construct a harness for each test task~\citep{qian2026learning}.
Meta$^2$ freezes reusable evolution guidance for edits during task execution, resetting the harness between task runs~\citep{yuan2026meta}.

\paragraph{Memory for task solving.}
Reflexion stores verbal reflections to guide later attempts~\citep{shinn2023reflexion}, and Voyager maintains a growing library of executable skills~\citep{wang2023voyager}.
ExpeL extracts reusable insights from task experience~\citep{zhao2024expel}; ReasoningBank and ACE retain strategies and lessons~\citep{ouyang2025reasoningbank,zhang2025agentic}.
ACE incrementally extends, refines, and deduplicates its playbook from execution trajectories.
Memory Transfer Learning compares task-experience representations across coding domains and models, finding abstract insights more transferable than concrete traces in its studied settings~\citep{kim2026memory}.
These memories support the agent executing tasks.

\paragraph{SIM in continual harness optimization.}
SIM supplies the proposer and selector with a record organized by harness problem, linking selected modifications to their rationale, predicted and observed outcomes, and evaluation context (\S\ref{sec:method-sim}).
The record accompanies the evolved harness across a benchmark or model transition and gains new attempts during target RSI.
Our comparisons with no SIM and root causes only test this complete procedure, combining inherited history with target-side updates (Appendix~\ref{sec:app-memory-ablation}).

\section{Benchmarks and Evaluation}
\label{sec:app-benchmarks}

This section specifies the protocol for the main cross-dataset and Opus-to-Gemini comparisons and the paired coding study.
We list the benchmarks and split sizes, explain which feedback can guide optimization, and then define the scoring rules for each domain.
The model-generation and shared-harness experiments have additional settings described in Appendices~\ref{sec:app-generations} and~\ref{sec:app-shared}, respectively.

\subsection{Which Benchmarks Do We Use?}
\label{sec:app-domains}

\paragraph{Benchmarks.}
We study four domains in which a source-evolved harness remains executable in the target environment.
For \emph{coding}, Terminal-Bench 2.1 (TB2.1) is the source and Terminal-Bench 4 (TB4) the target~\citep{merrill2026terminal,tbench2026v21,tbench2026v4}.
For \emph{text classification}, we use the online classification environment of Meta-Harness~\citep{lee2026meta}.
The source suite combines LawBench~\citep{fei2024lawbench}, the Gretel adaptation of Symptom2Disease~\citep{gretel2023symptom}, and USPTO-50k~\citep{schneider2016uspto}.
The target suite contains eight datasets absent from source RSI: AG News~\citep{zhang2015character}, GoEmotions~\citep{demszky2020goemotions}, Banking77~\citep{casanueva2020banking77}, Financial PhraseBank~\citep{malo2013good}, SciCite~\citep{cohan2019scicite}, TweetEval-Hate~\citep{barbieri2020tweeteval}, Amazon Reviews (Amazon5)~\citep{keung2020amazon}, and SciTail~\citep{khot2018scitail}.
For \emph{agentic workspace tasks}, Harvey LAB~\citep{harvey2026lab} is the source.
JobBench~\citep{li2026jobbench}, GDPval~\citep{patwardhan2026gdpval}, and APEX-Agents~\citep{vidgen2026apex} jointly form the target environment.
For \emph{engineering design}, EngDesign~\citep{guo2025engdesign} is the source and Frontier-Eng~\citep{chi2026frontiereng} the target.
Table~\ref{tab:benchmarks} gives the task inventory and rollout allocation. Appendix~\ref{sec:app-evaluation} defines the split and feedback-access rules; Appendices~\ref{sec:app-coding}--\ref{sec:app-engineering} detail each domain's scoring protocol.

\begin{table*}[t]
  \centering
  \resulttable
  \begin{tabular*}{\textwidth}{@{}ll@{\extracolsep{\fill}}rrrl@{}}
    \toprule
    \textbf{Domain} & \textbf{Benchmark} & \textbf{Total} & \shortstack[r]{\textbf{Evolve}\\tasks / rollouts} & \shortstack[r]{\textbf{Held-out}\\tasks / rollouts} & \textbf{Metric} \\
    \midrule
    \multirow{2}{*}{\textbf{Coding}}
      & S: TB2.1 & 89 & 44 / 220 & 45 / 225 & pass@1 \\
      & T: TB4 & 66 & 33 / 165 & 33 / 165 & pass@1 \\
    \midrule
    \multirow{2}{*}{\textbf{Text cls.}}
      & S: 3 datasets & 412 & 206 / 1,030 & 206 / 1,030 & accuracy \\
      & T: 8 datasets & 1,000 & 500 / 2,500 & 500 / 2,500 & accuracy \\
    \midrule
    \multirow{5}{*}{\textbf{Workspace}}
      & S: Harvey LAB & 160 & 80 / 400 & 80 / 400 & rubric \\
      & T: JobBench & 65 & 32 / 160 & 33 / 165 & rubric \\
      & T: GDPval & 220 & 110 / 550 & 110 / 550 & win rate \\
      & T: APEX-Agents & 480 & 240 / 1,200 & 240 / 1,200 & pass@1 \\
      & T: merged & 765 & 382 / 1,910 & 383 / 1,915 & mean of 3 \\
    \midrule
    \multirow{2}{*}{\textbf{Engineering}}
      & S: EngDesign & 67 & 34 / 170 & 33 / 165 & pass rate \\
      & T: Frontier-Eng & 47 & 24 / 120 & 23 / 115 & medal \\
    \bottomrule
  \end{tabular*}
  \caption{Benchmark inventory used in our experiments, with source (S) and target (T) roles for cross-dataset CRSI. Split sizes are tasks / rollouts (examples / rollouts for text classification). Each inventory is split uniformly at random into approximately equal halves, with five nominal rollouts per task. Frontier-Eng's overlapping EngDesign component receives no credit but remains in the denominator. The three workspace targets are evolved jointly and reported as their equal-weight mean; the merged row sums their task and rollout counts.}
  \label{tab:benchmarks}
\end{table*}

\paragraph{Source and target roles.}
In \emph{cross-dataset CRSI}, the policy model is fixed while the benchmark changes from the source to the target listed above.
In \emph{cross-model CRSI}, the benchmark and its task split are fixed while the policy changes: source RSI is run with the source policy on that benchmark, and target RSI continues with the target policy.
The Opus-to-Gemini comparisons use the corresponding T inventory in Table~\ref{tab:benchmarks} for both stages.

\subsection{How Are the Splits Used?}
\label{sec:app-evaluation}

The tasks (examples for text classification) in Table~\ref{tab:benchmarks} are split uniformly at random into approximately equal evolve and held-out halves, without stratification by category, difficulty, or source dataset.
When a benchmark holds an odd number of tasks, the evolve half takes the even count.
The two sets are disjoint, and no task or rollout is removed from the scoring denominator after the split.

Only evolve-set scores and execution trajectories are available for proposals, selection, SIM updates, and stopping decisions.
The task-solving policy executes the harness on held-out tasks to measure starting, final, and per-round diagnostic scores (\S\ref{sec:analysis}).
These held-out scores and trajectories are not passed to the harness-optimization agents.

Within these primary comparisons and the paired coding study, the scoring allocation is five rollouts per task on both splits, matched across conditions; the overlapping Frontier-Eng component contributes zero under the rule in Appendix~\ref{sec:app-engineering}.
The reported harness versions, independent-run counts, and statistical comparisons are specified in Appendix~\ref{sec:app-repetitions}.

\subsection{Coding: Terminal-Bench 2.1 and 4}
\label{sec:app-coding}

\paragraph{Environment and metric.}
Terminal-Bench tasks run in containers with a task description, shell access, and task-specific tests.
We report pass@1, estimated from the binary outcome of each task rollout.
For $n$ tasks with five rollouts each, let $y_{ir}\in\{0,1\}$ indicate whether rollout $r$ of task $i$ passes the task verifier.
The percentage score is
\begin{equation}
  \widehat{\mathrm{pass@1}}
  = \frac{100}{n}\sum_{i=1}^{n}\frac{1}{5}\sum_{r=1}^{5}y_{ir}.
  \label{eq:app-coding-pass1}
\end{equation}

\paragraph{Failed rollouts.}
A rollout that fails for infrastructure reasons stays in the denominator and scores zero.

\subsection{Text Classification}
\label{sec:app-classification}

The online classification environment follows Meta-Harness: the harness controls example storage, retrieval, and prediction context.
The example counts in Table~\ref{tab:benchmarks} are totals across each suite, not allocations for individual datasets.

\paragraph{Example memory during evaluation.}
The task agent reads only labeled examples supplied in advance from training/evolve data. Held-out labels are used for scoring and are not fed back into its example memory.
For a fixed harness, each of the five evaluation repetitions starts from the same initial example-memory state.
This memory supplies prediction context to the task agent and is separate from SIM, which supports harness optimization.

\paragraph{Accuracy across examples.}
We compute accuracy over all examples in a split, pooling the three source datasets or the eight target datasets.
For split $q\in\{\mathrm{evo},\mathrm{eval}\}$, let $n_q$ be the total number of examples and $y_{ir}(H;q)\in\{0,1\}$ indicate whether harness $H$ predicts the correct label for example $i$ in rollout $r$.
The percentage accuracy is
\begin{equation}
  A_{\mathrm{cls}}(H;q)=\frac{100}{5n_q}\sum_{i=1}^{n_q}\sum_{r=1}^{5}y_{ir}(H;q).
  \label{eq:app-classification-average}
\end{equation}
Each example has equal weight, so a dataset contributes in proportion to its example count.
The same aggregation is used on the evolve and evaluation splits.

\subsection{Agentic Workspace Tasks}
\label{sec:app-workspace}

\paragraph{Task scoring and judging.}
We adapt the benchmark-specific scoring interfaces described in RRSI~\citep[Appendix~B.3--B.6]{xia2026rrsi}. Our runs use Gemini 3.8 Flash as the judge for all workspace benchmarks; this differs from RRSI's mixed judge configurations and GDPval voting panel.
Harvey LAB reports the fraction of rubric criteria satisfied, and JobBench uses its weighted rubric score.
GDPval compares each generated deliverable with the benchmark's expert deliverable and reports the win rate against that reference.
APEX-Agents reports pass@1, with success determined by satisfaction of the task's rubric on a single rollout; the repeated rollouts estimate this success probability.
The judging interface and model are fixed across comparison conditions within each benchmark.
We express each benchmark score on a 0--100 scale before taking the equal-weight mean.

\paragraph{Equal-weight benchmark aggregation.}
For split $q\in\{\mathrm{evo},\mathrm{eval}\}$, let $S_b(H;q)$ be the benchmark-specific percentage score of harness $H$ on benchmark $b$.
The workspace score is
\begin{equation}
  S_{\mathrm{workspace}}(H;q)
  =\frac{1}{3}\sum_{b\in\mathcal{B}}S_b(H;q),
  \label{eq:app-workspace-aggregation}
\end{equation}
where $\mathcal{B}$ contains JobBench, GDPval, and APEX-Agents.
Each benchmark contributes one third, independent of its number of tasks or rubric criteria.
The evolve-set performance objective used to select target harnesses and the reported evaluation aggregate use this same weighting.
Within each benchmark, scoring retains that benchmark's own evaluation convention; averaging occurs after computing its score.
The workspace row of Table~\ref{tab:main-results} reports this aggregate.

\subsection{Engineering Design}
\label{sec:app-engineering}

Frontier-Eng has no MCP gateway of its own and uses the runner and gateway of EngDesign.
EngDesign uses a binary verifier pass; Frontier-Eng uses a medal score against frozen thresholds, with gold, silver, and bronze mapped to $1.00$, $0.67$, and $0.33$.
Neither engineering metric uses an LLM judge.

\paragraph{Source overlap.}
Frontier-Eng includes an EngDesign-derived component containing seven sub-problems~\citep{chi2026frontiereng}.
Following RRSI, we assign zero credit to this overlapping component while retaining it in the original denominator, including after the 24/23 split.

\section{Experimental Setup}
\label{sec:app-setup}

This section specifies the models, seed harnesses, comparison conditions, and optimization budgets.
The common setup covers cross-dataset and Opus-to-Gemini transfer (\S\ref{sec:results}--\ref{sec:results-model-transfer}); the last two subsections describe continuation across model generations and shared-harness construction (\S\ref{sec:results-generations}--\ref{sec:results-shared}).
Benchmark scoring is defined in Appendix~\ref{sec:app-benchmarks}, and SIM implementation in Appendix~\ref{sec:appendix}.

\subsection{Which Models and Budgets Are Used?}
\label{sec:app-implementation}

\paragraph{Task-solving policy.}
Gemini 3.8 Flash is the policy in both stages of cross-dataset CRSI.
The cross-family comparison uses Claude Opus 5.5 as the source policy and Gemini 3.8 Flash as the target policy.
Policy weights remain fixed.

\paragraph{Harness-optimization agents.}
The proposer and selector use Gemini 3.8 Flash, including when Opus executes the tasks.
The ablation in Appendix~\ref{sec:app-harness-agent} replaces the proposer and selector with Opus during source RSI.

\paragraph{Optimization budget.}
Each RSI run uses $N=10$ rounds with $k=5$ candidate proposals per round.
Candidate proposals and task rollouts are separate budgets: the main comparisons use five rollouts per task on each split (Appendix~\ref{sec:app-benchmarks}).
Proposals screened out before execution receive no rollout scores.

\paragraph{Cost per round.}
A coding round takes about 4\,h, including 1.7\,h for evaluation and 34\,min for candidate sampling.
We match rounds and candidate counts across methods; token use and runtime can differ.

\subsection{Which Harnesses Do We Start From?}
\label{sec:app-seeds}

We use publicly available seed harnesses from prior work. The two seeds in coding and text classification test different starting harnesses; repeated optimization runs are described in Appendix~\ref{sec:app-repetitions}.

\paragraph{Coding.}
The AHE seed starts with a basic shell tool and allows edits to prompts, tools, middleware, skills, sub-agents, and memory~\citep{lin2026agentic}.
Terminus-2 is distributed with Harbor and is also used by RRSI~\citep{merrill2026terminal,xia2026rrsi}.

\paragraph{Text classification.}
NoMemory stores no training examples; FewShotAll stores the observed labeled examples and includes them as few-shot context~\citep{lee2026meta}.

\paragraph{Workspace and engineering.}
We use RRSI's ReAct toolbelt agent for file manipulation, PDF reading, and code execution~\citep{yao2022react,vidgen2026apex,xia2026rrsi}.
Its 6 files (1,743 lines) are shared by the source and all three targets, allowing the Harvey LAB harness to run on the targets without modification.
Engineering uses the same agent with a bare system prompt and task wrapper, evaluated by fixed task-specific verifiers (Appendix~\ref{sec:app-engineering}).
This seed has 7 files (1,857 lines), with prompts in a separate module.
For these two domains, we adopt RRSI's environments, seed harnesses, and evaluation interfaces. Our models and budgets are specified above; our splits, rollout counts, and workspace judge are specified in Appendix~\ref{sec:app-benchmarks}.

\subsection{How Do the Four Methods Differ?}
\label{sec:app-transfer-conditions}

Figure~\ref{fig:transfer} summarizes the four conditions from \S\ref{sec:method-crsi}:
\begin{itemize}[leftmargin=*,nosep]
  \item \textbf{Cold Start}: the seed harness $H_0$, without SIM.
  \item \textbf{Harness Transfer}: the source-evolved harness $H_S$, without SIM.
  \item \textbf{Problem-Only Memory}: the same $H_S$ with root causes only, omitting modifications, outcomes, and provenance.
  \item \textbf{Full SIM}: the same $H_S$ with the complete source record $\mathcal{M}_S$, updated during target RSI.
\end{itemize}
Each condition persists throughout target RSI. All methods retain the ordinary target-run history $\mathcal{D}_t$.
All four use the same target policy, evolve tasks, round and candidate budgets, and evaluation allocation, and receive feedback through the same RSI loop.
Harness Transfer uses the evolved-harness starting point studied in prior work~\citep{lin2026agentic,lee2026meta,xu2026self}.

\begin{figure}[!ht]
  \centering
  \small
\setlength{\tabcolsep}{3pt}
\begin{tabularx}{\linewidth}{@{}l c X@{}}
\toprule
Condition & Harness & SIM during target RSI \\
\midrule
Cold Start & $H_0$ & None \\
Harness Transfer & $H_S$ & None \\
Problem-Only & $H_S$ & Root causes only \\
Full SIM & $H_S$ & Full record; updated on target \\
\bottomrule
\end{tabularx}

  \caption{Transfer conditions. The last three rows share the source-evolved harness. The listed memory condition persists throughout target RSI; the full record contains root causes, edits, outcomes, and provenance. Target round and candidate budgets match.}
  \label{fig:transfer}
\end{figure}

\subsection{How Are Runs Reported and Repeated?}
\label{sec:app-repetitions}

\paragraph{Reported harnesses.}
Each round retains one harness for the next round; $H_S$ is the final retained source harness.
The \emph{before} score evaluates the harness used to initialize target RSI, the \emph{after} score evaluates the version retained at the end of the final target round, and the target improvement is their difference.
Curves show each retained version's measured score, without taking a running maximum over earlier scores.

\paragraph{Independent runs.}
Repeated task rollouts and independent optimization runs are distinct sources of variation.
Each main cross-dataset and Opus-to-Gemini condition uses one complete RSI run.
The paired coding study repeats source RSI ten times from Terminus-2; each evolved harness initializes two target continuations, with and without SIM (\S\ref{sec:analysis}, Table~\ref{tab:paired}).
Task rollouts estimate execution performance. The ten independent source runs support across-run intervals for the paired study; scores from single-run conditions are reported descriptively.
For the paired study, the paired $t$ interval and two-sided test use the ten final-score differences; the signed-rank check uses average ranks for ties and enumerates the $2^{10}$ sign assignments.

\subsection{How Is Transfer across Generations Tested?}
\label{sec:app-generations}

We continue optimization through Gemini 2.5 Flash~\citep{gemini25}, 3.5 Flash~\citep{google2026gemini35}, 3.8 Flash~\citep{google2026gemini38}, and 4 Argon~\citep{google2026gemini4}, keeping TB2.1 as the benchmark.
Fresh RSI starts from the seed harness at each generation.
CRSI starts from the same seed in the first generation, then passes the evolved harness and SIM to each subsequent model.
The first row tests SIM during an initial RSI run; only the later rows test transfer from an earlier generation.

Table~\ref{tab:generations} reports scores to one decimal, with the results discussed in \S\ref{sec:results-generations}.

\paragraph{External scores use different evaluation settings.}
For context, the published score is 16.9 for Gemini 2.5 Flash with Terminus-2 on TB2.0~\citep{merrill2026terminal}, and 76.2 and 89.4 on TB2.1 for Gemini 3.5 and 3.8 Flash~\citep{google2026gemini35,google2026gemini38}.
These are external references, not matched controls. We include no external score for Gemini 4 Argon.

\subsection{How Is a Shared Harness Tested?}
\label{sec:app-shared}

We test a harness shared by Grok 4.20 Reasoning, Claude Opus 5.5, and Gemini 3.8 Flash on the text-classification target suite (Table~\ref{tab:multimodel}).
The evolve and held-out sets each contain 500 examples, with five rollouts per example (2,500 rollouts per split).

\paragraph{Harness construction.}
We compare two baselines and three ways to build a shared harness:
\begin{itemize}[leftmargin=*,nosep]
  \item \textbf{B1: Per-model harnesses.} Each policy model runs its own RSI and is evaluated with its own evolved harness.
  \item \textbf{B2: Shared seed.} All three models use the same unevolved harness.
  \item \textbf{S1: Sequential CRSI.} The order is Grok 4.20 Reasoning, Gemini 3.8 Flash, then Claude Opus 5.5, from weaker to stronger policies in this setting. Each model passes the evolved harness and SIM to the next.
  \item \textbf{S2: Unified synthesis.} Each round pools trajectory feedback from all three models to propose updates to one shared harness.
  \item \textbf{S3: Single-agent transfer.} We evolve a harness separately with each of the three policy models, then evaluate each resulting harness with all three models.
\end{itemize}
This experiment tests the performance of a common harness; Table~\ref{tab:crossmodel} separately tests the effect of transferring SIM from a matched starting harness.

\paragraph{Score aggregation.}
All averages over models use equal weights.
B1 averages the three model--harness pairs, each model using its own harness.
B2, S1, and S2 average the three models using a shared harness.
Each S3 row names the source policy used to evolve a harness and reports its mean score across the same three evaluation models.
The All column is $(\mathrm{Evolve}+\mathrm{Held\mbox{-}out})/2$, the average over the two equally sized splits.

Each procedure is run once, and optimization cost is not compared. The results in \S\ref{sec:results-shared} describe the observed means; small differences do not establish statistical equivalence or a ranking for each model.

\section{SIM Implementation}
\label{sec:appendix}

We supplement \S\ref{sec:method} with entry-matching details, prompt guidance, and the stored record format.

\subsection{How Are Entries Matched?}
\label{sec:app-sim-interfaces}

For the update in \S\ref{sec:method-sim}, the proposer itself decides whether a diagnosis matches an existing root cause; no separate similarity model is used.
Its choice is saved in the round's change manifest before evaluation.
Entries are merged deterministically when their root-cause text is identical, preserving all attempts.

\subsection{How Do the Proposer and Selector Use SIM?}
\label{sec:app-sim-rules}

The prompt guidance for \S\ref{sec:method-crsi} is summarized below:
\begin{enumerate}[leftmargin=*,nosep]
  \item Treat entries as prior experience, with flexibility to choose a different edit.
  \item Check the domain, benchmark, and model behind each entry.
  \item Start with an existing diagnosis when it explains the current failure.
  \item Reuse the rationale and reassess the edit after a benchmark or model change.
  \item Revise, replace, or override a harness modification when new evidence warrants it.
  \item Revisit a tested direction only when new evidence justifies it.
  \item Create a new problem entry when existing diagnoses do not explain the failure.
  \item Give task-specific evidence more weight in similar settings and less when both benchmark and model change.
\end{enumerate}

\subsection{What Does a SIM Record Contain?}
\label{sec:app-sim-format}

Each run stores SIM in one JSON file, with the following schematic structure:
\par\noindent\begin{minipage}{\linewidth}
\ifdefined\nolinenumbers\nolinenumbers\fi
\begin{lstlisting}[basicstyle=\ttfamily\scriptsize]
{ "next_id": 10,
  "carried_from": { "arm": <source run>, "domain": "coding_enriched",
                    "into": <target run> },
  "aliases": { "P6": "P5" },              // P6 was merged into P5
  "problems": [ ...entries P1 P2 P3 P4 P5 P7 P8 P9... ] }
\end{lstlisting}
\end{minipage}\par
Each problem entry contains \texttt{\{id, name, opened\_round, attempts[]\}}.
\texttt{name} gives the root-cause hypothesis, and \texttt{opened\_round} records its first diagnosis.
Each attempt contains:
\texttt{\{round, sample, accepted, backfilled, what, files, predicted, fixed, missed, passed, total, domain\}}.
\texttt{what} and \texttt{files} describe the edit, and \texttt{domain} identifies its setting.

\texttt{predicted} lists the tasks expected to be fixed before evaluation; \texttt{fixed} and \texttt{missed} record the outcomes afterward.
\texttt{passed}/\texttt{total} gives the successful and total trials in that attempt's evolve-set evaluation.
\texttt{accepted} means the selector chose the edit in that round, even if some predicted fixes failed.
The edit may later be changed or replaced while its record stays in SIM, as in the P8 case in Appendix~\ref{sec:app-case-gating}.
Predictions for the same task in different rounds are separate observations.

\texttt{carried\_from} identifies the source and target runs; \texttt{aliases} keeps the identifiers of merged entries (for example, P6$\rightarrow$P5).
Read each entry's ID and \texttt{opened\_round} within its recorded run context.
Appendix~\ref{sec:app-cases} gives records as stored, including P3 in Figure~\ref{fig:sim-p3}.

\section{Additional Results}
\label{sec:app-results}

We report source-RSI scores (\S\ref{sec:app-source}), exact target scores (\S\ref{sec:app-bars}), source retention (\S\ref{sec:app-retention}), and memory and harness-agent ablations (\S\ref{sec:app-ablations}).
The model-generation, shared-harness, and ten-pair coding results are in the main text (Tables~\ref{tab:generations}, \ref{tab:multimodel}, and \ref{tab:paired}); edit histories and cases are in Appendix~\ref{sec:app-cases}.

\label{sec:app-diagnostic-provenance}
Evaluation and reporting follow Appendices~\ref{sec:app-benchmarks} and~\ref{sec:app-repetitions}; differences are absolute score points.

\subsection{Does Source RSI Improve the Seed Harness?}
\label{sec:app-source}

\begin{figure*}[!t]
  \centering
  \input{figures/source_results_include}
  \caption{Source RSI with Gemini 3.8 Flash: scores of the seed harness $H_0$ and the source-evolved harness $H_S$ on the source evolve set and the held-out source evaluation set. (a)~Coding on TB2.1 from the AHE seed; (b)~coding on TB2.1 from Terminus-2; (c)~text classification on the source suite from NoMemory; (d)~the same from FewShotAll; (e)~workspace tasks on Harvey LAB; (f)~engineering design on EngDesign. $H_S$ is the starting harness of Harness Transfer, Problem-Only Memory, and Full SIM in Figure~\ref{fig:main}. Open and teal bars show seed and source-evolved scores for each split, with both values labeled. All panels share a zero baseline and the same score scale.}
  \label{fig:source}
\end{figure*}

\paragraph{Source RSI improves all six held-out source scores.}
Source RSI improves the held-out source score in all six settings, by 2.75--6.41 points (Figure~\ref{fig:source}).
In five settings, the held-out gain exceeds the evolve-set gain; the exception is FewShotAll, with held-out and evolve-set gains of 6.41 and 6.60 points.
Section~\ref{sec:results} examines whether these source gains yield a better target starting point.

\subsection{How Do the Methods Compare on the Target?}
\label{sec:app-bars}

Tables~\ref{tab:main-results} and~\ref{tab:crossmodel} give the numerical results plotted in Figures~\ref{fig:main} and~\ref{fig:crossmodel}, respectively.

\begin{table*}[!t]
  \centering
  \maintable
  \begin{tabular*}{\textwidth}{@{}l@{\extracolsep{\fill}}rr rr rr rr@{}}
    \toprule
    & \multicolumn{2}{c}{Cold Start} & \multicolumn{2}{c}{Harness Transfer} & \multicolumn{2}{c}{Problem-Only Memory} & \multicolumn{2}{c}{\textbf{Full SIM (ours)}} \\
    \cmidrule(lr){2-3}\cmidrule(lr){4-5}\cmidrule(lr){6-7}\cmidrule(l){8-9}
    \textbf{Domain / seed} & Before & After & Before & After & Before & After & Before & After \\
    \midrule
    Coding / AHE & 16.36 & 19.39 & 18.18 & 20.61 & 18.18 & 18.79 & 18.18 & \textbf{24.85} \\
    Coding / Terminus-2 & 18.18 & 19.39 & 19.39 & 19.39 & 19.39 & 20.61 & 19.39 & \textbf{21.82} \\
    \addlinespace[3pt]
    Text cls. / NoMemory & 68.60 & 72.00 & 71.40 & 71.00 & 71.40 & 73.40 & 71.40 & \textbf{73.60} \\
    Text cls. / FewShotAll & 75.72 & 75.92 & 72.92 & 74.12 & 72.92 & 75.52 & 72.92 & \textbf{77.92} \\
    \addlinespace[3pt]
    Workspace / ReAct & 73.89 & 82.98 & 74.93 & 85.07 & 74.93 & 85.54 & 74.93 & \textbf{87.89} \\
    Engineering / ReAct & 39.13 & 40.87 & 39.13 & 42.61 & 39.13 & 42.61 & 39.13 & \textbf{46.96} \\
    \bottomrule
  \end{tabular*}
  \caption{Cross-dataset CRSI: held-out target scores before and after target RSI for the six settings of Figure~\ref{fig:main}. The first column identifies the domain and seed harness. Bold marks the highest final score in each setting.}
  \label{tab:main-results}
\end{table*}

\begin{table*}[!t]
  \centering
  \maintable
  \begin{tabular*}{\textwidth}{@{}l@{\extracolsep{\fill}}rr rr rr rr@{}}
    \toprule
    & \multicolumn{2}{c}{Cold Start} & \multicolumn{2}{c}{Harness Transfer} & \multicolumn{2}{c}{Problem-Only Memory} & \multicolumn{2}{c}{\textbf{Full SIM (ours)}} \\
    \cmidrule(lr){2-3}\cmidrule(lr){4-5}\cmidrule(lr){6-7}\cmidrule(l){8-9}
    \textbf{Domain / seed} & Before & After & Before & After & Before & After & Before & After \\
    \midrule
    Coding / AHE & 16.36 & 19.39 & 13.94 & 18.18 & 13.94 & 20.61 & 13.94 & \textbf{21.82} \\
    Text cls. / FewShotAll & 78.32 & 82.72 & 77.72 & 80.20 & 77.72 & 81.32 & 77.72 & \textbf{83.80} \\
    Workspace / ReAct & 73.89 & 82.98 & 68.41 & 73.42 & 68.41 & 79.22 & 68.41 & \textbf{84.75} \\
    Engineering / ReAct & 39.13 & 40.87 & 34.78 & 42.61 & 34.78 & 46.96 & 34.78 & \textbf{50.43} \\
    \bottomrule
  \end{tabular*}
  \caption{Cross-model CRSI (Claude Opus 5.5 to Gemini 3.8 Flash): exact held-out scores for Figure~\ref{fig:crossmodel}. Before and After refer to Gemini's target RSI. Both stages use the same benchmark and split; the three transfer conditions share the Opus-evolved starting harness. The first column identifies the domain and seed harness. The Terminus-2 Cold Start counterpart for coding scores 18.18 before and 19.39 after target RSI. Bold marks the highest final score in each row.}
  \label{tab:crossmodel}
\end{table*}

\subsection{How Much Source Performance Is Retained?}
\label{sec:app-retention}

\begin{figure}[!t]
  \centering
  \input{figures/forgetting_include}
  \caption{Source scores during target RSI (coding, TB2.1$\rightarrow$TB4, Gemini 3.8 Flash): held-out TB2.1 score (45 tasks, 225 rollouts; one rollout is 0.44 points) of the harness retained after each target round from the AHE seed, with one run per condition. The three transfer conditions start from the source-evolved harness; Cold Start begins from the seed.}
  \label{fig:forgetting}
\end{figure}

\paragraph{Both memory conditions retain more source performance in this run.}
In Figure~\ref{fig:forgetting}, Full SIM and Problem-Only Memory end at 88.89 and 88.44, respectively, from the common starting score of 90.22.
Their lowest source score is 88.44, a drop of 1.78 points.
Harness Transfer falls to 86.67 in round 2 and ends at 87.56; Cold Start begins at 85.78 and ends at 85.33.
Section~\ref{sec:analysis} relates this trace to target progress: the two memory conditions retain similar source performance, while Full SIM improves more on TB4.

\subsection{What Do the Ablations Show?}
\label{sec:app-ablations}

\subsubsection{Is Recording Only Root Causes Enough?}
\label{sec:app-memory-ablation}

From the same evolved harness, Full SIM finishes above both no-memory and root-cause-only continuation in all ten settings (Tables~\ref{tab:main-results} and~\ref{tab:crossmodel}).
This comparison tests each memory condition throughout target RSI, covering both inherited information and target-stage updates (Appendix~\ref{sec:app-transfer-conditions}).
The main results discuss the margins and the mixed effects of retaining root causes alone (\S\ref{sec:results} and \S\ref{sec:results-model-transfer}).

\subsubsection{Does the Choice of Harness Agent Matter?}
\label{sec:app-harness-agent}

\begin{table*}[!t]
  \centering
  \resulttable
  \begin{tabular*}{\textwidth}{@{}ll@{\extracolsep{\fill}}rrrrr@{}}
    \toprule
    & & & \multicolumn{2}{c}{\textbf{Gemini 3.8 Flash}} & \multicolumn{2}{c}{\textbf{Claude Opus 5.5}} \\
    \cmidrule(lr){4-5}\cmidrule(l){6-7}
    \textbf{Set} & \textbf{Seed} & \textbf{Before} & After & Gain & After & Gain \\
    \midrule
    Evolve & AHE        & 86.82 & 90.45 & +3.63 & \textbf{91.23} & +4.41 \\
           & Terminus-2 & 85.91 & 89.09 & +3.18 & \textbf{92.21} & +6.30 \\
    \addlinespace[5pt]
    Held-out & AHE        & 85.78 & 90.22 & +4.44 & \textbf{91.28} & +5.50 \\
             & Terminus-2 & 87.56 & 91.56 & +4.00 & \textbf{91.61} & +4.05 \\
    \bottomrule
  \end{tabular*}
  \caption{Source RSI on TB2.1 with Gemini 3.8 Flash or Claude Opus 5.5 as the harness agent, the optimization agent that proposes and selects harness edits: the score of the seed harness, and the score after source RSI with each harness agent. Gain is the change from Before; bold marks the higher After score within each row.}
  \label{tab:harness-agent}
\end{table*}

\paragraph{Opus scores higher, with a smaller held-out margin on Terminus-2.}
Table~\ref{tab:harness-agent} replaces Gemini 3.8 Flash with Claude Opus 5.5 as the proposer and selector during TB2.1 source RSI.
The held-out score is 1.06 points higher with Opus for AHE and 0.05 points higher for Terminus-2.
This experiment varies the optimization agents; the policy-model transitions are studied separately in \S\ref{sec:results-crossmodel}.

\section{Case Studies}
\label{sec:app-cases}

We distinguish improvements within a fixed setting from continued optimization after transfer.
The source-RSI cases show changes to command transport and parsing (\S\ref{sec:app-case-source}).
The CRSI cases follow completion control from TB2.1 to TB4 (\S\ref{sec:app-case-gating}), compare continuations with and without SIM (\S\ref{sec:app-component}), and trace new evidence appended to P3 (\S\ref{sec:app-case-p3}).
These are complementary views of the coding transfer study, not independent repetitions.
We then summarize transfer patterns in the four domains (\S\ref{sec:app-transfer}) and draw together the lessons for SIM (\S\ref{sec:app-case-lessons}).
The detailed coding cases use Terminus-2 and Gemini 3.8 Flash.

\paragraph{Evidence in the cases.}
\label{sec:app-case-evidence}
These cases document edit histories rather than estimate each edit's contribution to the final score.
Code fragments are simplified; displayed SIM records retain their stored text, including truncations.
Appendix~\ref{sec:app-sim-format} defines the fields.

\subsection{How Does RSI Improve the Source Harness?}
\label{sec:app-case-source}

These examples stay within source RSI: feedback identifies a harness failure and leads to an edit, without a benchmark or policy change.
The transport case includes a measured task outcome; the parser example documents the repair mechanism.

\subsubsection{\texttt{adaptive-rejection-sampler}: unreliable command transport}
\textbf{Task.} Implement adaptive rejection sampling in R as \texttt{/app/ars.R}, with input validation and tests.

\textbf{Root cause ($c$).} The agent wrote the whole program in a single shell command of about 22.8\,KB, exceeding the command-transport limit reported in this run (about 16\,KB). The command never ran, so the file was never created and the verifier failed with \texttt{No ars.R file found}.

\textbf{Modification ($a$).} The transport behind the existing \texttt{run\_shell\_command} tool now splits literal input into chunks of at most 2{,}048 characters, sends special keys such as \texttt{Enter} as separate key events, and raises an error when a send fails instead of failing silently (simplified):
\par\noindent\begin{minipage}{\linewidth}
\ifdefined\nolinenumbers\nolinenumbers\fi
\begin{lstlisting}[language=Python]
MAX_KEY_CHUNK_SIZE = 2048
def _tmux_send_keys(self, keys):
    cmds = []
    for key in keys:
        if key in SPECIAL_KEYS:  # key event
            cmds.append(f"{prefix} {quote(key)}")
            continue
        for chunk in self._split_key(key):
            cmds.append(f"{prefix} {quote(chunk)}")
    return cmds
\end{lstlisting}
\end{minipage}\par

\textbf{Outcome ($o$).} Successful rollouts increased from 1 to 5 out of 5 on this task.

\subsubsection{\texttt{gpt2-codegolf}: fragile tool-call parsing}
\textbf{Task.} Write a compact GPT-2 inference program in C (\texttt{/app/gpt2.c}) that loads the weights, decodes greedily, and compiles with \texttt{gcc -O3 -lm} under a code-size limit.

\textbf{Root cause ($c$).} Terminus-2 expects a JSON response whose \texttt{keystrokes} fields hold shell input. When the model wrote C code containing unescaped quotation marks into that field, for example
\par\noindent\begin{minipage}{\linewidth}
\ifdefined\nolinenumbers\nolinenumbers\fi
\begin{lstlisting}
"keystrokes": "...FILE *f = fopen("model.bin", "rb");..."
\end{lstlisting}
\end{minipage}\par
\texttt{json.loads} failed and the command was rejected. The parser could already add missing closing braces and strip surrounding text, but not repair quotes inside a field.

\textbf{Modification ($a$).} A repair step, run only after normal parsing fails, locates the real closing quote of each string field as the one followed by the next expected key (for \texttt{keystrokes}, \texttt{"duration"} or the closing brace) and escapes every quote before it.

\textbf{Outcome ($o$).} The change adds a recovery path for malformed calls; no post-edit task success rate is reported for this parser repair.
Its dependence on the model's output format is discussed in \S\ref{sec:app-transfer-coding}.
The P3 record in \S\ref{sec:app-case-p3} concerns a separate prompt intervention.

\subsection{How Does CRSI Adapt an Evolved Harness?}
\label{sec:app-case-gating}

This case follows completion control across a benchmark change, with the policy fixed.
Source RSI builds an execution check on TB2.1; target RSI inherits it and adds checks for solution quality on TB4.
P5 records the source problem of premature completion; P8 adds a target diagnosis about solution quality.
Both concern the inherited completion gate, but they are separate entries, with no explicit parent/refinement link.
\subsubsection{Source: Execution-Verified Completion}

\paragraph{Task and evaluation interface.}
\emph{``Configure a git server so that from my own machine I can \texttt{git clone user@server:/git/server}, commit \texttt{hello.html}, push, and then \texttt{curl http://server:8080/hello.html} shows `hello world'.''}
The agent must build four parts that compose: a bare repository at \texttt{/git/server}, a \texttt{post-receive} hook that deploys to the web root, an SSH-reachable account named \texttt{user}, and an HTTP service on port 8080.
The verifier runs outside the container and clones as \texttt{user} over SSH, so a setup that works locally is not evidence that the externally visible interface is correct.

\paragraph{Observed failure and diagnosis.}
In the described failing rollout, the agent configured the repository, hook, and web server, read \emph{``I will set up login myself''} as meaning the account would be provisioned externally, and never created \texttt{user}.
It tested the flow with the pre-existing \texttt{ubuntu} account, saw every local command succeed, and declared \texttt{task\_complete: true}.
SSH then refused the nonexistent user, the push failed, and the page returned 404.
The seed harness could not catch this, because its completion check only asks the model to repeat its claim (simplified from \texttt{terminus\_2.py}):
\par\noindent\begin{minipage}{\linewidth}
\ifdefined\nolinenumbers\nolinenumbers\fi
\begin{lstlisting}[language=Python]
if is_task_complete:
    if self._pending_completion:   # second claim
        return episode + 1         # terminate
    self._pending_completion = True  # first: ask again
    obs = self._confirmation_message(out)
\end{lstlisting}
\end{minipage}\par
and the confirmation message asks about intent, not evidence: \emph{``Are you sure you want to mark the task as complete? \ldots\ If so, include \texttt{"task\_complete": true} in your JSON response again.''}
\textbf{Root cause:} ending the episode is an agent privilege that requires no execution evidence.

\paragraph{Harness intervention.}
The intervention introduces completion-control middleware in the harness.
Source RSI built it over three accepted rounds:

\begin{center}
\footnotesize
\renewcommand{\arraystretch}{1.18}
\setlength{\tabcolsep}{4pt}
\begin{tabularx}{\linewidth}{@{}lY@{}}
\toprule
\textbf{Round} & \textbf{Edit} \\
\midrule
r6 & New \texttt{execution\_gating\_interlock.py} (137 lines), wired into \texttt{terminus\_2.py}: a cross-turn state machine that refuses completion when the previous turn had a parse error or no command actually executed, capped at three rejections. \\
r8 & Also intercepts simulated multi-turn text, truncated output with discarded command blocks, and completion after at most one real command (adds \texttt{terminus\_json\_plain\_parser.py}). \\
r10 & Detects terminal crashes, assertion failures, HTTP 404s, and test errors at the moment completion is requested (adds both prompt templates). \\
\bottomrule
\end{tabularx}
\end{center}

\noindent The interlock tracks execution across turns and uses four rejection paths (simplified):
\par\noindent\begin{minipage}{\linewidth}
\ifdefined\nolinenumbers\nolinenumbers\fi
\begin{lstlisting}[language=Python]
def check_completion_attempt(self, cmds, out, pending=False):
    if self._consecutive_rejections >= self._max_rej:
        return False, ""  # never lock out a done task
    if self.detect_terminal_error(out):      # Case 0
        return self._reject(...)
    if pending:
        return False, ""
    ran = any(c.keystrokes.strip() for c in cmds)
    if self._last_turn_had_error and not ran:   # A
        return self._reject(...)
    if self._last_turn_discarded and not ran:   # B
        return self._reject(...)
    if self._total_executed <= 1 and not ran:   # C
        return self._reject(...)
    return False, ""
\end{lstlisting}
\end{minipage}\par
The interlock enforces the completion checks; prompt reminders provide additional guidance.
By the end of source RSI the file had grown to 275 lines.

\paragraph{Reported task outcome.}
Successful rollouts increased from 1 to 5 out of 5 on \texttt{configure-git-webserver}.
The same P5 intervention also raised \texttt{circuit-fibsqrt} from 1 to 5 out of 5: that task tests Fibonacci and square-root circuits with \texttt{/app/sim.c}, including large-integer inputs, and the agent had stopped after failed commands.
These are two task outcomes for the same source intervention.

\begin{figure*}[!t]
\centering
\begin{minipage}{\textwidth}
\ifdefined\nolinenumbers\nolinenumbers\fi
\begin{lstlisting}[basicstyle=\ttfamily\footnotesize]
{ "id": "P5",
  "name": "premature completion without execution verification",
  "opened_round": 6,
  "attempts": [
    { "round": 6, "sample": "s2", "accepted": true, "domain": "coding_enriched",
      "what": "Implemented ExecutionGatingInterlock in Terminus2 to intercept premature
               task completion attempts when previous commands failed to execute
               (e.g. on unescaped JSON parse errors) or when no commands wer...",
      "files": "execution_gating_interlock.py, terminus_2.py",
      "predicted": ["circuit-fibsqrt", "configure-git-webserver"],
      "fixed":     ["circuit-fibsqrt", "configure-git-webserver"],
      "missed": [], "passed": 49, "total": 60 },
    { "round": 8, "sample": "s1", "accepted": true, "domain": "coding_enriched",
      "what": "Intercept premature completion when LLM generates multi-turn simulated
               text or output-length truncations with unexecuted discarded command
               blocks or completes with trivial session execution (<=1 comma...",
      "files": "execution_gating_interlock.py, terminus_2.py,
                terminus_json_plain_parser.py  +1 more",
      "predicted": ["constraints-scheduling"],
      "fixed":     ["constraints-scheduling"],
      "missed": [], "passed": 45, "total": 60 },
    { "round": 10, "sample": "s1", "accepted": true, "domain": "coding_enriched",
      "what": "Enhance ExecutionGatingInterlock in Terminus2 to detect active terminal
               crashes, assertion failures, HTTP 404s, and test errors when task
               completion is requested, prevent immediate termination when co...",
      "files": "execution_gating_interlock.py, templates/terminus-json-plain.txt,
                templates/terminus-xml-plain.txt  +1 more",
      "predicted": ["configure-git-webserver", "protein-assembly", "train-fasttext"],
      "fixed":     ["configure-git-webserver"],
      "missed":    ["protein-assembly", "train-fasttext"],
      "passed": 47, "total": 60 } ] }
\end{lstlisting}
\end{minipage}
  \caption{SIM entry P5 as carried into the target run's record (\texttt{what} fields truncated as stored). All three attempts are source rounds, identified by \texttt{domain: "coding\_enriched"}.}
  \label{fig:sim-p5}
\end{figure*}

The three source attempts remain together in P5 (Figure~\ref{fig:sim-p5}).
Two target proposals cited P5, but neither was selected, so no target attempt was appended to that entry.

\subsubsection{Target: Feasibility and Solution Quality}

\paragraph{Task and evaluation interface.}
Route nine photonic waveguides described in \texttt{/app/layout\_spec.json} and write the waypoint paths to \texttt{/app/routing\_result\_1.json}.
Each path must run from its exact start to its exact target coordinate using 90-degree bends of radius $50\,\mu$m and valid s-bends, stay inside the board, avoid obstacle clearance regions and self-intersections, and keep a minimum separation between nets.
The score is the negative weighted route cost summed over all nets.
The task ships its own feasibility checker, \texttt{check\_routing.py}, which answers whether the constraints are satisfied, while the reward is a cost function.
A solution can pass every check and still score poorly.

\paragraph{Observed failure and diagnosis.}
The evolved harness contains the source interlock, intact and wired.
On this task the agent produces a feasible routing, writes the output, runs \texttt{check\_routing.py}, sees it pass, encounters no terminal error, and requests completion.
None of the four rejection conditions is triggered: commands ran, no parse error occurred, nothing was discarded, and no visible terminal failure was detected, so the gate allows completion.
The target failure adds a question about sufficient evidence for completion:

\begin{center}
\footnotesize
\renewcommand{\arraystretch}{1.18}
\setlength{\tabcolsep}{4pt}
\begin{tabularx}{\linewidth}{@{}>{\raggedright\arraybackslash}p{0.24\linewidth}YY@{}}
\toprule
& \textbf{The source gate asks} & \textbf{This task also needs} \\
\midrule
Question & Has enough executed for the workflow not to have obviously failed? & Is the feasible solution good enough to stop improving? \\
Evidence & Execution activity, errors, crashes & Objective value against an achievable bound \\
\bottomrule
\end{tabularx}
\end{center}

\paragraph{Harness intervention and installed behavior.}
Target RSI retained the state machine and bounded rejection, expanding the component from 275 to 437 lines under several target entries.
The change attributed to this deficiency, accepted in target round 2, adds a detector for the signature of a constraint-only check that also prints a score, and turns a match into a notice at the confirmation step (simplified):
\noindent\begin{minipage}{\linewidth}
\begin{lstlisting}[language=Python]
PATTERNS = [re.compile(
    r"(?:ALL CHECKS PASSED|ASSERTIONS PASSED"
    r"|CONSTRAINTS? SATISFIED).*?"
    r"(?:score|cost|loss|metric)\s*[:=]\s*[-+]?\d+",
    re.I | re.S), ...]

def detect_optimization_feasibility_notice(cls, out):
    if any(p.search(out[-4000:]) for p in PATTERNS):
        return ("[ExecutionGatingInterlock NOTICE]: "
                "passing constraint checks (e.g. "
                "check_routing.py) only proves FEASIBILITY, "
                "NOT OPTIMALITY ... compare your current "
                "score against theoretical bounds ...")
\end{lstlisting}
\end{minipage}

The pre-completion checklist gained a matching clause asking the agent to verify that it ``has not stopped at the first feasible solution'' and has ``quantified [its] suboptimality gap against the theoretical bound.''
In target round 4, the adopted candidate added a fifth rejection path, blocking completion once when feasibility output with a metric was detected.
Later edits changed this behavior: the final harness retains the source body of \texttt{check\_completion\_attempt} and its four rejection paths.
The error patterns feeding Case~0 were expanded to catch deliverables that report themselves infeasible (e.g.\ \texttt{"route\_feasible": false}).
The final behavior is therefore to block reported infeasibility and warn about possible suboptimality; SIM retains the earlier blocking attempt and its outcome.

\paragraph{Recorded task outcome.}
The recorded predictions and outcomes distinguish the intended repair from its measured result:

\begin{center}
\footnotesize
\renewcommand{\arraystretch}{1.18}
\setlength{\tabcolsep}{4pt}
\begin{tabularx}{\linewidth}{@{}lYl@{}}
\toprule
\textbf{Attempt} & \textbf{Predicted} & \textbf{Fixed} \\
\midrule
P8, round 2 & \texttt{photonic-waveguide-routing}, \texttt{production-planning} & none \\
P8, round 4 & \texttt{photonic-waveguide-routing}, \texttt{production-planning} & none \\
\bottomrule
\end{tabularx}
\end{center}

\noindent Both attempts were selected, but neither fixed its targeted tasks.
Their predictions remain marked as unresolved in P8 (Figure~\ref{fig:sim-p8}).
This case illustrates adaptation and memory updates; aggregate target performance is reported separately in \S\ref{sec:results}.

\begin{figure*}[!t]
\centering
\begin{minipage}{\textwidth}
\ifdefined\nolinenumbers\nolinenumbers\fi
\begin{lstlisting}[basicstyle=\ttfamily\footnotesize]
{ "id": "P8",
  "name": "premature termination on feasible but suboptimal solutions
           in optimization tasks",
  "opened_round": 2,
  "attempts": [
    { "round": 2, "sample": "s3", "accepted": true, "domain": "tb4",
      "what": "Add objective optimality vs constraint feasibility guidelines to prompt
               templates and enhance ExecutionGatingInterlock with optimality audit
               checklist and feasibility score notice to prevent premature...",
      "files": "execution_gating_interlock.py, templates/terminus-json-plain.txt,
                templates/terminus-xml-plain.txt  +1 more",
      "predicted": ["photonic-waveguide-routing", "production-planning"],
      "fixed": [],
      "missed":    ["photonic-waveguide-routing", "production-planning"],
      "passed": 5, "total": 33 },
    { "round": 4, "sample": "s4", "accepted": true, "domain": "tb4",
      "what": "Active optimization gating and terminal error precision in
               ExecutionGatingInterlock: enforce completion interception when
               constraint feasibility outputs with score/metric are detected
               without verified...",
      "files": "execution_gating_interlock.py",
      "predicted": ["photonic-waveguide-routing", "production-planning"],
      "fixed": [],
      "missed":    ["photonic-waveguide-routing", "production-planning"],
      "passed": 10, "total": 36 } ] }
\end{lstlisting}
\end{minipage}
  \caption{SIM entry P8, opened by target RSI in round 2 (\texttt{what} fields truncated as stored). Its provenance, \texttt{domain: "tb4"}, places it inside the target run. The round-4 modification was adopted and later changed; its record remains in SIM.}
  \label{fig:sim-p8}
\end{figure*}

The two entries distinguish the inherited execution problem from the new solution-quality problem:

\begin{center}
\footnotesize
\renewcommand{\arraystretch}{1.18}
\setlength{\tabcolsep}{4pt}
\begin{tabularx}{\linewidth}{@{}>{\raggedright\arraybackslash}p{0.24\linewidth}YY@{}}
\toprule
& \textbf{P5 (carried)} & \textbf{P8 (target-native)} \\
\midrule
Provenance & \texttt{coding\_enriched}, source round 6 & \texttt{tb4}, target round 2 \\
Missing evidence & Execution did not occur or did not succeed & Execution succeeded but the solution is only feasible \\
Marked accepted & 3 of 3, all on the source & 2 of 2, both on the target \\
Realized predictions & 4 of 6 & 0 of 4 \\
Diagnosis opened in & Source RSI & Target RSI \\
Component & Interlock created & Interlock extended \\
Completion behavior & Four source rejection paths & Quality advice in final harness \\
\bottomrule
\end{tabularx}
\end{center}

\paragraph{Memory preserves the history as the harness changes.}
P5 preserves why the source interlock was introduced; P8 records the target attempts to assess solution quality, including the blocking attempt that was later changed.
Together they retain the rationale and outcomes beyond the edits present in the final harness.

\subsection{How Does SIM Guide Continuation?}
\label{sec:app-component}

\begin{figure*}[!t]
  \centering
  \begin{minipage}[b]{\textwidth}
\centering
\includegraphics[width=0.94\textwidth]{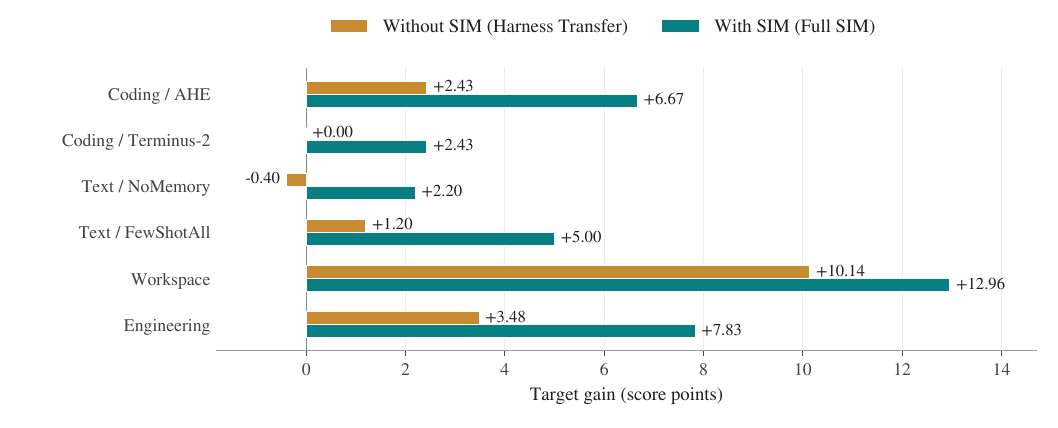}\\[2pt]
{\small\sffamily (a) Target gain from the same inherited harness}
\end{minipage}\par\vspace{10pt}
\begin{minipage}[b]{\textwidth}
\centering
\footnotesize
\setlength{\tabcolsep}{3pt}
\renewcommand{\arraystretch}{1.12}
\begin{tabular}{@{}c p{0.43\linewidth} p{0.43\linewidth}@{}}
  \toprule
  Round & \textbf{With SIM}: same gate + source record & \textbf{Without SIM}: same gate, no record \\
  \midrule
  R1 & Gate + specification-parameter audit & Gate: stronger error patterns \\
  R2 & Gate + feasible $\neq$ optimal check & Prompt: configuration completeness \\
  R3 & tmux transport fix & Gate: fix completion deadlocks \\
  R4 & Gate + feasibility-score block (P8) & --- \\
  R5 & Gate + coupled-field and test-suite audit & Gate: cross-turn test-failure latching \\
  R6 & Gate + defensive function contracts & --- \\
  R7 & Gate + infeasibility flags & Gate: fix latched-error deadlock \\
  R8 & --- & --- \\
  R9 & Gate + lifecycle-state check & --- \\
  \bottomrule
\end{tabular}\\[4pt]
{\small\sffamily (b) Retained edits during target RSI}
\end{minipage}

  \caption{Target gains and edit histories with and without SIM. (a)~The six cross-dataset target gains from Figure~\ref{fig:main}, measured from the common source-evolved starting harness. (b)~Selected edits in target rounds 1--9 from Terminus-2 (TB2.1$\rightarrow$TB4). Both runs inherit the same harness and completion gate. ``Gate +'' adds a check; --- marks no selected edit; round 10 is not listed. The round-4 P8 block was later changed (\S\ref{sec:app-case-gating}); rounds 5 and 6 extend P3 (Figure~\ref{fig:sim-p3}).}
  \label{fig:with-without}
\end{figure*}

\paragraph{The SIM run adds target checks to the inherited gate.}
Figure~\ref{fig:with-without}b follows the completion gate inherited from TB2.1.
It compares the continuation described in \S\ref{sec:app-case-gating} with a run from the same harness without SIM.
In round 2, the proposer with SIM reads the gate's source entry, P5, which diagnoses premature completion without execution verification.
The proposer keeps the gate and asks the agent to compare its score with theoretical bounds before declaring completion: satisfying constraints establishes feasibility, but a better solution may still exist.
Later edits add checks of specification parameters and lifecycle state.
Without SIM, the run edits the same gate four times, twice to repair completion deadlocks.
Its round-3 rationale attributes the gate and the deadlocks to round 1, although the gate was inherited from source RSI.

Appendix~\ref{sec:app-case-gating} gives the task-level evidence and memory records behind these revisions.

\subsection{How Does SIM Extend an Existing Record?}
\label{sec:app-case-p3}

\begin{figure*}[!t]
  \centering
\begin{minipage}{\linewidth}
\small
\noindent{\color{crsiTeal}\rule{\linewidth}{0.8pt}}\par
\vspace{3pt}
\textbf{SIM entry P3}\hfill Policy: Gemini 3.8 Flash\par
\vspace{3pt}
\textbf{Root cause.} A deliverable can be internally correct yet violate the verifier's external contract: schema, coupled attributes, API signature, or split hygiene.\par
\vspace{6pt}
\renewcommand{\arraystretch}{1.16}
\setlength{\tabcolsep}{5pt}
\begin{tabularx}{\linewidth}{@{}>{\raggedright\arraybackslash}p{0.15\linewidth} >{\raggedright\arraybackslash}X >{\raggedright\arraybackslash}p{0.30\linewidth}@{}}
\toprule
\textbf{Stage} & \textbf{Selected modification} & \textbf{Recorded outcome} \\
\midrule
\textcolor{cPO}{\textbf{Source}}\newline TB2.1, round 4\newline Carried forward
& \emph{System prompt.} Execution and Verification Guidelines: deliverable integrity, schema conformance, defensive API compatibility, evaluation hygiene, and invariant checks instead of silent suppression.
& \textbf{1 of 4 fixed.} Fixed \texttt{gpt2-codegolf}; \texttt{protein-assembly}, \texttt{torch-pipeline-parallelism}, and \texttt{train-fasttext} remained unresolved. \\
\addlinespace[5pt]
\textcolor{crsiTeal}{\textbf{Target}}\newline TB4, round 5\newline Appended
& \emph{Verification layer.} Add a schema-deliverables clause to the completion interlock and prompt templates; extend the checklist to programmatic assertions and a test-suite audit.
& \textbf{0 of 2 fixed.} Neither \texttt{atrx-vep-crispr} nor \texttt{intrastat-meldung} was fixed. \\
\addlinespace[5pt]
\textcolor{crsiTeal}{\textbf{Target}}\newline TB4, round 6\newline Appended
& \emph{Verification layer.} Add a defensive-contract detector to the completion checklist.
& \textbf{1 of 2 fixed.} Fixed \texttt{embedding-drift-monitor}; \texttt{biped-contact-dynamics} remained unresolved. \\
\bottomrule
\end{tabularx}
\end{minipage}

  \caption{SIM entry P3 across source and target RSI. The source attempt realizes one of four task-fix predictions; target rounds 5 and 6 realize zero of two and one of two. All attempts retain their outcomes and provenance under the same root-cause hypothesis, including unresolved predictions.}
  \label{fig:sim-p3}
\end{figure*}

Figure~\ref{fig:sim-p3} shows the record behind the round-5 and round-6 gate edits in Figure~\ref{fig:with-without}b.
Source RSI diagnoses a mismatch between internally correct deliverables and the external schema or API contract checked by the verifier.
The round-4 source edit adds system-prompt guidance and fixes one of four predicted tasks, leaving three unresolved.
Target RSI appends two attempts under the same diagnosis: a schema-deliverables and test-audit extension in round 5, and a defensive-contract detector in round 6.
They fix zero of two and one of two predicted tasks, respectively.
P3 therefore retains one diagnosis while accumulating evidence from different interventions and settings.

\subsection{What Transfers across Datasets and Models?}
\label{sec:app-transfer}

Table~\ref{tab:app-transfer} summarizes which mechanisms are reused, adapted, or extended, and which depend on the target setting.
The domain summaries distinguish recorded changes from conditions that limit reuse; a retained mechanism can remain inactive.
They broaden the coding cases to the other domains without treating every mechanism as a separate SIM ablation.

\begin{table*}[tp]
  \centering
  \footnotesize
  \setlength{\tabcolsep}{3pt}
  \renewcommand{\arraystretch}{1.22}
  \begin{tabularx}{\textwidth}{@{}>{\raggedright\arraybackslash}p{1.85cm}YYYY@{}}
    \toprule
    \textbf{Domain / shift} & \textbf{Reused directly} & \textbf{Reused with adaptation} & \textbf{Inherited and extended} & \textbf{Setting-dependent} \\
    \midrule
    \textbf{Coding}\newline Dataset & Shell, terminal, and process layer & Paste-buffer transport (key normalization, foreground detection) & Completion contract, with new target predicates & Source completion predicates; refusal handling (idle) \\
    Model & Terminal and process layer; interlock accounting & --- & Lifecycle control, with recalibrated interception heuristics & Parser repairs tuned to one model's output \\
    \midrule
    \textbf{Text cls.}\newline Dataset & BM25 index and retriever; query sanitization & Top-$k$, score filter, and class quotas (recalibrated constants) & Multi-template routing skeleton & Data-specific post-processing (SMILES normalization) \\
    Model & Indexing; query sanitization & Retrieval budgets (re-tuned to model capacity) & Routing skeleton, with rewritten rubrics & Output schema and decoding contract \\
    \midrule
    \textbf{Workspace}\newline Dataset & Soft gate before \texttt{final\_answer} & Evidence probe (file system $\rightarrow$ invocation history) & Truncation limits and context budget & Evaluator-specific registration assumption; substring heuristics \\
    Model & System-prompt injection pipeline & Budget-triggered sparse intervention & Rubric taxonomy with counterexamples & Calibrations to one model's failure rate or API schema \\
    \midrule
    \textbf{Engineering}\newline Dataset & Central tool-call interception point & Deliverable check $\rightarrow$ continuous checkpointing & Context buffers and truncation budgets & Prompt guidelines without SIM (reverted) \\
    Model & Deliverable observation layer; re-execution check & Intervention schedules (time and step thresholds) & Composable soft-gate predicates & Intervention intensity; environment-specific bypasses \\
    \bottomrule
  \end{tabularx}
  \caption{Mechanism reuse across datasets and models. The last column includes inactive or removed components as well as assumptions and heuristics that require reassessment in a new setting. A dash means that no mechanism is listed. \S\ref{sec:app-transfer-coding}--\S\ref{sec:app-transfer-eng} give the supporting cases.}
  \label{tab:app-transfer}
\end{table*}

\subsubsection{Coding}
\label{sec:app-transfer-coding}

\paragraph{Source RSI revises execution, parsing, and completion.}
The evolved tool layer retains stdout and stderr after timeouts, checks background processes and their exit codes, recovers misplaced artifacts, and sends long commands through a paste buffer (\S\ref{sec:app-case-source}).
Terminus-2's parser attempts to repair malformed JSON before rejecting it and detects safety refusals.
Completion control requires evidence before accepting a completion claim: AHE's middleware forces continuation when the agent stops without a tool call, and Terminus-2's cross-turn interlock blocks completion after unexecuted commands or errors (Appendix~\ref{sec:app-case-gating}).

\paragraph{Core execution handling is retained on TB4.}
The command-execution routines carry over with the shared container and terminal interfaces, while session checks adapt as described below.
The recorded cold-start run does not rediscover the tool-layer changes within eight rounds; after ten rounds it has a larger parser without outperforming the inherited harness.

\paragraph{Paste-buffer checks adapt to TB4.}
TB4 exposes failures absent from TB2.1: special key sequences must be normalized for interrupts to take effect, and foreground processes must be identified so that a running command is not classified as a stalled session.
The paste-buffer structure remains, with checks adjusted to these target failures.

\paragraph{Completion checks add target criteria.}
Both inherited harnesses keep completion control while revising the evidence required on TB4.
Appendix~\ref{sec:app-component} follows the selected edits; Appendix~\ref{sec:app-case-gating} gives the task evidence and the P5/P8 records.
Source predicates that no longer fire are pruned rather than re-tuned.
Both branches also add a reminder triggered by the remaining step budget in their first round, a target-specific mechanism that the source did not need.

\paragraph{Refusal handling remains inactive.}
Refusal handling is carried forward but has no triggering tasks on TB4.
Its presence alone therefore does not establish a transfer benefit or a loss.

\paragraph{Parser repairs depend on the policy model.}
The Opus- and Gemini-evolved harnesses have near-identical terminal and process structures, which are retained after transfer.
Parser repairs instead address model-specific output errors: unescaped control characters, illegal escapes, and truncated responses.
Each repair needs checking against the target model's outputs.
Completion control likewise retains counts of executed commands, discarded blocks, and consecutive rejections, while recalibrating checks for simulated execution and premature completion.
The target run also adds format constraints and step-by-step guidance absent from source RSI.

\subsubsection{Text Classification}
\label{sec:app-transfer-cls}

\paragraph{Source RSI builds structured retrieval.}
Memory writing moves from no retention (NoMemory) or unstructured storage (FewShotAll) to retrievable indices: multi-granularity BM25 over words, character $n$-grams, and chemical-formula $n$-grams, later extended with class-based candidate pools, hard-negative error indices, and class-prototype centrality.
Prompt assembly bounds retrieval to the top eight examples with positive BM25 scores, strips benchmark boilerplate from queries before retrieval, and selects contrastive, class-balanced exemplars.
Post-processing normalizes data-specific formats, such as SMILES strings for USPTO-50k, because open-vocabulary generation requires exact matches.

\paragraph{Retrieval survives with adjusted thresholds.}
The target suite changes the input types and label spaces.
The BM25 index and retriever classes remain intact, while retrieval constants are recalibrated and query sanitization gains delimiters for the new datasets.
The routing structure is retained, with rubrics and exemplars rewritten for the target.
With SIM, target RSI adds smoothed class quotas and contrastive ordering; class balancing applies to the target's finite label sets rather than the source's open-vocabulary outputs.

\paragraph{SMILES normalization stays inactive.}
SMILES normalization used in source RSI remains inactive on the target suite, whose outputs no longer use that format.
The cold-start run independently develops relevance-gated retrieval and bounded context.

\paragraph{Indexing transfers while output rules change.}
Three of four runs evolve BM25 indices from different starting harnesses, using features suited to the input, such as chemical $n$-grams for SMILES and character bigrams for Chinese legal text.
Indexing and query sanitization depend on the input data and carry over across models; retrieval budgets such as top-$k$ are re-tuned.
Output rules depend more on policy behavior: the answer-first schema addresses Gemini's truncated reasoning.
Target RSI also rewrites inherited rubrics more extensively on Banking77 and GoEmotions than on SciTail and TweetEval-Hate.

\subsubsection{Agentic Workspace}
\label{sec:app-transfer-ws}

\paragraph{Source RSI adds output handling and audits.}
Over-long tool outputs are preserved through head-and-tail truncation, with quotas raised three- and five-fold, respectively.
A one-time audit gate before \texttt{final\_answer} can append a tool message and return the policy to the loop.
A deliverable auditor supplies the gate with file names and sizes from candidate output directories.

\paragraph{The termination gate survives a rubric change.}
Source RSI and a cold-start target run independently add a tool message before the same early-return branch, with different predicates and messages.
The target gate fires only when no deliverable is registered, files were written during code execution, and the gate has not fired before; the second condition keeps purely conversational tasks from being intercepted.
This insertion point depends on the ReAct loop and remains usable when the evaluation rubric changes.

\paragraph{Registration checks replace file-system probes.}
The source probe samples the file system, whereas target failures come from invocation history: files created on disk but never registered through \texttt{write\_file}.
Target checks therefore use tool-invocation history rather than file-system probes.
The target escalates this check in three single-shot tiers: an advisory notice when file saving is detected, a mandatory instruction once 75\% of the step budget passes without registration, and a conditional block at \texttt{final\_answer}.

\paragraph{Registration checks depend on the evaluator.}
The registration check assumes that the evaluator credits files registered through \texttt{write\_file}; it would need reconsideration for an evaluator that reads output directories directly.
Its substring matches over tool arguments (such as \texttt{/tmp/}, \texttt{.docx}, or \texttt{savefig}) also depend on argument formatting.

\paragraph{Prompt infrastructure transfers across models.}
The Opus source run adds editable system-prompt guidelines and a six-level severity taxonomy with counterexamples, leaving control flow and tool execution unchanged.
The injection mechanism uses message structures, while the taxonomy describes the unchanged evaluator.
Gemini's cold-start run takes a different direction: a notice at 80\% of the step budget asks it to fill pending fields so that a deliverable exists before \texttt{final\_answer}; seven later rounds add no improvement in this trace.
The timing and wording of such interventions depend on the policy's behavior.

\subsubsection{Engineering Design}
\label{sec:app-transfer-eng}

\paragraph{Source RSI adds design and verification guidance.}
The source run adds prompt guidance on analytical calculations, self-contained deliverables, tool-failure recovery, and re-reading written files before completion.
It also imports a syntax parser and a tool registry without calling them during source RSI.

\paragraph{Target RSI expands checks and context budgets.}
The tool-registry import is unused in the source harness but becomes the entry point that the Harness Transfer target run uses to enforce centralized output truncation at the registry call.
Three target runs independently raise the shared context constants: output-token caps, head and tail lengths, and summarization triggers.
The Full SIM target run adapts the source's one-time post-generation check into a checkpoint protocol that writes intermediate artifacts exceeding the baseline score to disk, protecting them from unannounced timeouts.
The cold-start target run's retained edits are limited to prompt changes: reconstructing the objective from the evaluation script, embedding self-tuning optimization code in deliverables, and clamping outputs with a fallback to the baseline implementation.

\paragraph{SIM retains the source guidelines in this trace.}
The Harness Transfer target run inherits the source guidelines but reverts them to the seed prompt in its first round, whereas the Full SIM target run retains and extends them throughout.
This pair retains similar code structures but differs in whether the inherited guidance is preserved.

\paragraph{Deliverable checks survive model changes.}
A deliverable tracker that records deliverable state independently of model output transfers without refactoring, and a check that blocks \texttt{final\_answer} when a deliverable was modified but not re-executed transfers unchanged.
Budget prompts emerge as wall-clock triggers in an Opus run (every 600\,s, at most three times) and step-budget ratios in Gemini runs (0.8 and 0.75).
Both use the gate before \texttt{final\_answer}, with checks on different state variables.
The recorded zero-score rates differ between Opus (0.09\%) and Gemini (4.49\%); intervention thresholds therefore need checking for each policy.
Handlers for a missing sandbox binary remain idle where the environment is fully provisioned.

\paragraph{Interfaces explain much of the observed reuse.}
\label{sec:app-transfer-patterns}
Across these cases, benchmark changes alter the evidence needed for completion, while model changes alter how the harness handles policy failures.
With the benchmark fixed, a model change leaves the success criterion intact but can require different detection and guidance.
Stable interfaces favor reuse; output repairs and intervention thresholds need checking against the target setting.

\subsection{What Do the Cases Tell Us about SIM?}
\label{sec:app-case-lessons}

The cases suggest four roles for SIM in continuing harness optimization.

\paragraph{Earlier rationale informs what to retain.}
In engineering, the SIM run retains and extends inherited guidance, while harness-only continuation reverts it (\S\ref{sec:app-transfer-eng}).
Recording why an edit was made gives the optimizer a basis for assessing its continued relevance, beyond its presence in the inherited harness.

\paragraph{A transferable rationale can require a different implementation.}
The coding runs inherit the same completion gate, but the SIM run extends it to check target-specific requirements (\S\ref{sec:app-component}).
The need for completion evidence carries over; what counts as sufficient evidence changes.
This illustrates how earlier reasoning can guide adaptation without treating the source implementation as fixed.

\paragraph{New outcomes add evidence to an existing diagnosis.}
P3 links a source diagnosis to further target interventions and their outcomes (\S\ref{sec:app-case-p3}).
Keeping these attempts together lets later rounds inspect which remedies were tried, under which settings, and which predictions remain unresolved.
The reusable experience includes the evidence behind a diagnosis, not only its name.

\paragraph{Experience remains available after an edit is replaced.}
P8 retains the selected blocking attempt and its unresolved predictions after later edits change the behavior (\S\ref{sec:app-case-gating}).
The final harness alone would omit this part of the search history.
SIM keeps the attempt, its diagnosis, and its unresolved outcome together as the harness changes.

\end{document}